\documentclass{article} %
\usepackage{utt,times}

\usepackage{amsmath,amsfonts,bm}

\def\eqref#1{equation~\ref{#1}}

\def\1{\bm{1}}

\DeclareMathAlphabet{\mathsfit}{\encodingdefault}{\sfdefault}{m}{sl}
\SetMathAlphabet{\mathsfit}{bold}{\encodingdefault}{\sfdefault}{bx}{n}

\usepackage{graphicx}
\usepackage{multirow}
\usepackage{tabularx}
\usepackage{amsmath,amssymb}
\usepackage{marvosym}

\usepackage{algorithm}
\usepackage{algpseudocode}

\usepackage{hyperref}
\usepackage{url}
\usepackage{wrapfig}
\usepackage{float}
\usepackage{booktabs}
\usepackage{enumitem}
\usepackage{tablefootnote}
\usepackage{threeparttable}
\usepackage{xcolor}

\usepackage{amsmath,amssymb,tabularx,fvextra,enumitem,tcolorbox}
\tcbuselibrary{breakable,skins,listings}

\newcommand{\UTThead}[1]{%
  \par\medskip\noindent\textbf{#1}\par\nobreak\smallskip
}

\newtcolorbox[auto counter]{UTTPrimerBox}[2][]{
  enhanced,
  breakable,
  colback=black!1,
  colframe=black!35,
  colbacktitle=black!6,
  coltitle=black,
  boxrule=0.4pt,
  arc=1pt,
  left=6pt,
  right=6pt,
  top=5pt,
  bottom=5pt,
  before skip=8pt,
  after skip=10pt,
  fonttitle=\small\bfseries,
  fontupper=\small,
  title={Primer Example~\thetcbcounter: #2},
  title after break={Primer Example~\thetcbcounter\space(continued)},
  before upper={
    \setlength{\parindent}{0pt}
    \setlength{\parskip}{3pt}
    \setlength{\emergencystretch}{2em}
    \setlist[itemize]{
      leftmargin=1.4em,
      itemsep=2pt,
      parsep=0pt,
      topsep=3pt
    }
    \setlist[enumerate]{
      leftmargin=1.8em,
      itemsep=2pt,
      parsep=0pt,
      topsep=3pt
    }
    \raggedright
  },
  #1
}

\newtcblisting{UTTRolePromptBox}[1]{
  enhanced,
  breakable,
  colback=black!1,
  colframe=black!35,
  colbacktitle=black!6,
  coltitle=black,
  boxrule=0.4pt,
  arc=1pt,
  left=6pt,
  right=6pt,
  top=5pt,
  bottom=5pt,
  before skip=8pt,
  after skip=10pt,
  fonttitle=\small\bfseries,
  title={#1},
  title after break={#1 (continued)},
  listing only,
  listing options={
  basicstyle=\ttfamily\footnotesize,
  columns=fullflexible,
  keepspaces=false,
  showstringspaces=false,
  breaklines=true,
  breakatwhitespace=true,
  breakautoindent=false,
  breakindent=0pt
}
}

\newenvironment{UTTrow}
  {\par\noindent
   \tabularx{\linewidth}
   {@{}>{\raggedright\arraybackslash}X
       >{\raggedright\arraybackslash}X@{}}}
  {\endtabularx\par}

\title{Universal Textual Teaching for LLMs} 

\author{Zhanyi Lu \\
Westlake University \\
\texttt{luzhanyi@sjtu.edu.cn}
\And
Huan Wang\,\textsuperscript{\Letter} \\
Westlake University \\
\texttt{wanghuan@westlake.edu.cn}
}
\iclrfinalcopy
\hypersetup{
  pdftitle={Universal Textual Teaching for LLMs},
  pdfauthor={Zhanyi Lu and Huan Wang}
}
\usepackage{etoolbox}
\makeatletter
\patchcmd{\@maketitle}{\vskip 0.3in minus 0.1in}{\vskip 5pt}{}{}
\makeatother
\begin{document}
\setlength{\parskip}{5pt}

\maketitle
\lhead{UTT, Z. Lu and H. Wang}
{\centering\small\url{https://alexlu99.github.io/UTT/}\par}
\begingroup
\setlength{\intextsep}{5pt plus 1pt minus 1pt}
\setlength{\abovecaptionskip}{4pt}
\begin{figure*}[h]
    \centering
    \includegraphics[width=1\textwidth]{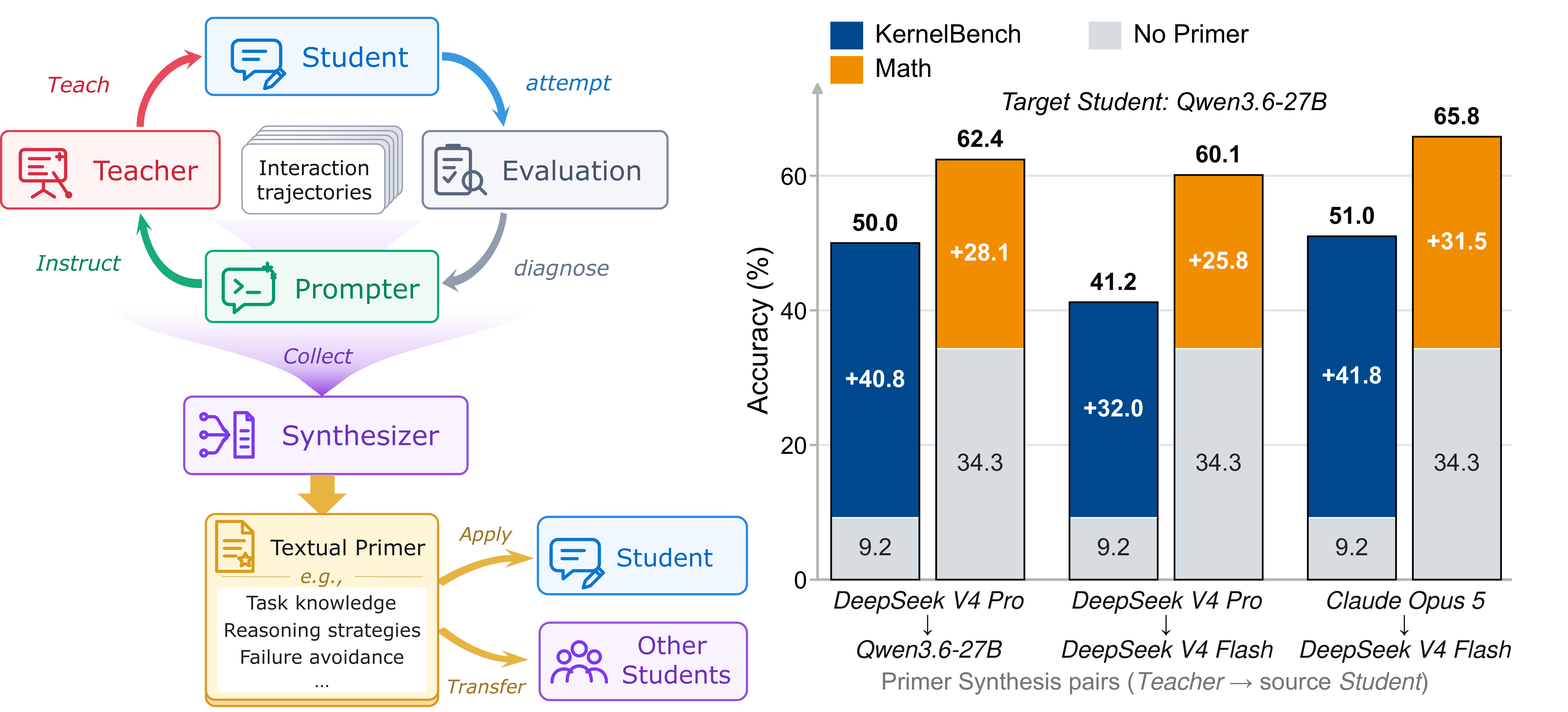}
\caption{
\textbf{Left:} We introduce \textit{\underline{U}niversal \underline{T}extual \underline{T}eaching} (UTT), a parameter-update-free framework that distills an LLM's knowledge into a textual \emph{Primer} through multi-role LLM interaction. Notably, the resulting \emph{Primer} can improve the performance of the source \textit{Student} and other \textit{Students} that are not involved in creating the \textit{Primer}. 
\textbf{Right:} For the target \textit{Student} Qwen3.6-27B, \textit{Primers} synthesized from the \textit{Teacher}-\textit{Student} pairs (shown at the bottom of the figure) consistently improve accuracy. Even when Qwen3.6-27B does not participate in the \textit{Primer} synthesis, the absolute accuracy gains reach 41.8\% and 31.5\% (see the rightmost column) on KernelBench and Omni-MATH-2, respectively.
These results suggest that text can serve as a universal medium for knowledge transfer among different LLMs for complex tasks.
}
    \label{fig:UTT-teaser}
\end{figure*}
\endgroup

\begingroup
\setlength{\topsep}{2pt}
\begin{abstract}

Knowledge distillation (KD) transfers knowledge from stronger \textit{Teacher} models to weaker \textit{Student} models, but most methods require training the \textit{Student} parameters, thereby binding the distilled knowledge to a specific architecture and checkpoint.
This implicit representation is difficult to interpret or reuse across models and limits KD for API-only or costly-to-train models.
This paper studies knowledge transfer for large language models (LLMs). We introduce \textbf{Universal Textual Teaching} (UTT), a parameter-update-free framework that distills observed \textit{Teacher-Student} knowledge gaps into a textual, interpretable, and reusable natural-language artifact called \emph{Primer}.
Specifically, UTT first identifies representative gap cases through paired evaluations, and iteratively updates the \emph{Primer} via multi-role interactions: the \textit{Student} attempts each task, the \textit{Prompter} turns evaluation feedback into a teaching instruction, the \textit{Teacher} provides a targeted demonstration, and the \textit{Synthesizer} consolidates validated lessons.
Empirically,
on the challenging math (Omni-MATH-2) and code generation (KernelBench) tasks, extensive results confirm the effectiveness of the method: UTT remarkably raises the Student's accuracy from $9.4\%$ to $48.6\%$ and Fast$_1$ accuracy from $9\%$ to $35\%$ on KernelBench, while increasing mathematical reasoning accuracy from $27.6\%$ to $51.7\%$.
UTT also performs better than representative prompt engineering and parameter-based KD methods.
Of note, UTT is shown to be \textit{generalizable} across different \textit{Teachers} and \textit{Students}: a \emph{Primer} synthesized for one \textit{Teacher}-\textit{Student} pair can generalize to other \textit{Students} that do not participate in the synthesis.
\end{abstract}
\endgroup
\setlength{\parskip}{6pt}

\section{Introduction}
Large language models (LLMs) demonstrate strong capabilities in reasoning, code generation, and knowledge-intensive tasks, yet capability gaps remain across models.\citep{gpt4, gemini, deepseekv4, qwen36_27b, chatglm}
Although open-source models are more flexible and accessible, they often trail powerful proprietary models because of limited scale and training compute \citep{scalinglaws}.
Knowledge distillation (KD) \citep{hinton2015distilling} transfers knowledge from large teacher models (\textit{Teachers}) to more efficient student models (\textit{Students}), reducing computational and energy costs, broadening access to advanced model capabilities, and enabling wider participation in AI research and development \citep{llm_kd_survey_2024}.

Existing LLM distillation methods typically follow two routes: supervised fine-tuning on teacher-generated answers or reasoning traces \citep{seqkd, distilling-step-by-step, Fine-tune-CoT}, and preference optimization or reinforcement learning guided by teacher evaluations \citep{minillm, gkd, luffy}.
Although effective, most require training the \textit{Student}, binding the distilled knowledge to a specific architecture and checkpoint.
Such implicit parameterization makes knowledge difficult to inspect or transfer, and limits applicability to API-only, non-trainable, or costly-to-train models.

These limitations raise a central question: \emph{Can the Teacher-\textit{Student} knowledge gap be distilled into an explicit, interpretable, and reusable natural-language representation without updating parameters?}
We build on three premises: (1) modern LLMs can understand and follow natural-language instructions; (2) task-specific gaps may reflect missing knowledge or operational principles; and (3) such knowledge, when verbalized, can be directly used by other LLMs.
Achieving this goes beyond asking a \textit{Teacher} to generate a general prompt from a task description: reliable transfer must identify representative problems solved by the \textit{Teacher} but not the \textit{Student}, extract reusable knowledge, and avoid degrading the \textit{Student}'s existing capabilities.

To this end, this paper presents \textbf{Universal Textual Teaching} (UTT), a parameter-update-free framework that distills model knowledge into a natural-language artifact called \textit{Primer}\footnote{We adopt the term \emph{Primer} in its original sense of an introductory textbook, reflecting the foundational instruction and guidance that a \textit{Teacher} provides to the \textit{Student}.} through multi-role teaching interactions (Figure~\ref{fig:UTT-teaser}).
UTT first identifies representative knowledge-gap cases.
For each case, the \textit{Student} attempts the problem, the \textit{Prompter} diagnoses the gap from evaluation feedback and formulates a teaching instruction, the \textit{Teacher} provides a demonstration, and the \textit{Synthesizer} integrates the record into the evolving global \textit{Primer}.
The \textit{Primer} guides subsequent teaching rounds and is deployed directly as natural-language context after distillation.
Relying solely on interactions among LLMs, the process requires no model training or parameter updates and supports both open-source and API-only commercial models.

Empirically, across various \textit{Teacher}-\textit{Student} configurations in mathematical reasoning and code generation, the synthesized \textit{Primers} consistently improve their source \textit{Students} and remain effective when transferred to other \textit{Students} without resynthesis, demonstrating both effectiveness and cross-model reusability.
Relative to the unprimed \textit{Student} baselines, UTT improves KernelBench accuracy from $9.4\%$ to $48.6\%$, Fast$_1$ from $9\%$ to $35\%$, and mathematical reasoning accuracy from $27.6\%$ to $51.7\%$.
Across the evaluated settings, UTT also outperforms representative prompt engineering and parameter-updating knowledge distillation methods.

Our main contributions are:

\textbf{(1) A language-based perspective on KD for LLMs.}
We formulate LLM distillation as representing the \textit{Teacher}-\textit{Student} knowledge gap in natural language rather than encoding it in a specific \textit{Student}'s parameters, making the distilled knowledge interpretable and reusable across models.

\textbf{(2) A parameter-update-free closed-loop KD framework.}
UTT identifies knowledge gaps through paired evaluation and iteratively synthesizes a deployable global \textit{Primer} from \textit{Student} attempts, \textit{Prompter} feedback, and \textit{Teacher} demonstrations, without updating any model parameters.

\textbf{(3) Empirical validation across tasks and models.}
Across math and code tasks, UTT demonstrates effectiveness and transferability, with \textit{Primers} transferable across models without resynthesis. UTT also outperforms representative prompt engineering and parameter-updating KD methods.

\section{Related Work}
\textbf{Knowledge distillation for LLMs.}
Knowledge distillation aims to transfer knowledge from a stronger \textit{Teacher} to a \textit{Student}, with existing methods broadly following two routes \citep{hinton2015distilling,llm_kd_survey_2024}.
One primarily relies on supervised learning, where the \textit{Teacher} provides examples such as answers, reasoning traces, or explanations, and the \textit{Student} learns by fitting these examples.
Representative methods include SeqKD \citep{seqkd}, which distills complete output sequences; Fine-tune-CoT \citep{Fine-tune-CoT} and Distilling Step-by-Step \citep{distilling-step-by-step}, which incorporate reasoning; and RSR \citep{rsr}, which constructs supervision informed by \textit{Student} performance.
The other emphasizes on-policy distillation and reinforcement learning, incorporating \textit{Student}-generated trajectories into optimization and learning from \textit{Teacher} feedback or task rewards.
For example, MiniLLM \citep{minillm} optimizes reverse KL divergence through policy gradients, GKD \citep{gkd} performs distribution matching on \textit{Student}-generated sequences, and LUFFY \citep{luffy} combines \textit{Teacher} demonstrations with \textit{Student} trajectories for reinforcement learning.

Despite differences in learning and supervision, these methods typically store transferred knowledge by updating \textit{Student} parameters, binding the distilled knowledge to specific model weights.
This limits independent inspection, editing, and cross-model reuse of the knowledge and requires trainable \textit{Student} parameters.
UTT retains the objective of knowledge distillation while explicitly representing the \textit{Teacher's} knowledge advantage over the \textit{Student} as a natural-language \textit{Primer}, which enhances the \textit{Student} through context while keeping its parameters fixed.

\textbf{Prompting techniques and engineering.}
Prompting techniques improve LLM performance through reusable input structures or invocation workflows, while prompt engineering selects, combines, evaluates, and refines these techniques \citep{prompt_report_2024}.
Representative techniques include in-context learning \citep{gpt3}, which uses examples for task adaptation, and Chain-of-Thought \citep{cot}, which elicits intermediate reasoning steps.
Automatic prompt optimization (APO) aims to reduce reliance on manual design \citep{ape_survey_2025}.
APE \citep{ape} searches model-generated candidate instructions; ProTeGi \citep{protegi} and OPRO \citep{opro} use textual feedback or candidate histories to guide optimization; MIPROv2 \citep{mipro} jointly optimizes instructions and demonstrations; and TextGrad \citep{textgrad} and GEPA \citep{gepa} use textual feedback and reflection to refine prompts or compound LLM systems.

UTT shares with APO the use of natural-language text optimization to enhance models, but differs in its research objective and organization of supervision.
APO searches and refines prompts directly for target-task performance, without using the knowledge gap between a designated \textit{Teacher} and \textit{Student} as its basis for optimization.
Motivated by knowledge distillation, UTT explicitly targets the Teacher's observed knowledge advantage over the \textit{Student} for transfer, organizing it into an inspectable \textit{Primer} reusable across \textit{Students}.

\textbf{Textual knowledge representations.}
Recent work has also explored natural-language representations of models and task knowledge.
Verbalized Machine Learning \citep{xiao2025vml} learns natural-language model parameters, while Explaining Datasets in Words \citep{zhong2024nlparams} uses natural-language predicates to construct statistical explanations of data; both primarily address predictive modeling or dataset explanation.
ExpeL \citep{zhao2024expel}, Dynamic Cheatsheet \citep{suzgun2025dynamic}, and SkillGLoW \citep{yan2026skillglow} accumulate reusable experience from task trajectories, inference-time memory, and execution-based validation, respectively.
AutoManual \citep{chen2024automanual} constructs instruction manuals and SkillX \citep{wang2026skillx} builds hierarchical skill libraries, also demonstrating the potential of knowledge constructed by stronger models to enhance weaker ones.
Their construction procedures primarily focus on environmental rules or general-purpose skills, without using the \textit{Teacher}-\textit{Student} performance gap as their basis.

UTT pursues knowledge distillation by identifying what to transfer from the paired performance and synthesizing a global \textit{Primer} from \textit{Student} attempts, evaluation feedback, and validated \textit{Teacher} demonstrations.
Its distinction lies in using the observed \textit{Teacher}-\textit{Student} knowledge gap to guide teaching-content selection and text synthesis, while examining whether the resulting knowledge transfers to other \textit{Students} without resynthesis.

\section{Universal Textual Teaching}
\label{sec:method}

This section introduces UTT, a parameter-update-free framework that distills \textit{Teacher}-\textit{Student} knowledge gaps into natural-language representations.
We first reformulate KD as optimizing a natural-language knowledge carrier rather than parameters, and then instantiate this formulation through knowledge-gap construction and multi-role interaction.

\subsection{From Parametric to Textual Knowledge Distillation}
\label{sec:problem_formulation}
Given a task distribution $\mathcal{D}$, a \textit{Teacher} $M_{\theta_T}$, and a \textit{Student} $M_{\theta_S}$, let $\theta_T\in\mathbb{R}^{d_T}$ and $\theta_S\in\mathbb{R}^{d_S}$ denote their parameters.
Conventional KD fixes $\theta_T$ and optimizes $\theta_S$ to learn from the \textit{Teacher}.
Parametric knowledge distillation is formulated as

\begin{equation}
\theta_S^{*}
=
\arg\min_{\theta_S\in\mathbb{R}^{d_S}}
\mathbb{E}_{x\sim\mathcal{D}}
\left[
\mathcal{F}_{\mathrm{KD}}
\left(
M_{\theta_S}(x),
\mathcal{K}_{M_{\theta_T}}(x)
\right)
\right],
\qquad
\theta_T\ \text{fixed}.
\label{eq:parametric-kd}
\end{equation}
For $x\sim\mathcal{D}$, $\mathcal{K}_{M_{\theta_T}}(x)$ denotes the instructional signal provided by the \textit{Teacher}, and $\mathcal{F}_{\mathrm{KD}}$ is the scalar objective used to train the \textit{Student} from this signal.
In supervised KD, the instructional signal typically comprises \textit{Teacher's} answers or reasoning traces, and the objective is commonly negative log-likelihood or cross-entropy; on-policy distillation methods instead incorporate \textit{Student}'s trajectories and optimize reinforcement learning objectives.
Despite differences in instructional signals and optimization objectives, these methods all optimize $\theta_S$ and encode the distilled knowledge in the updated \textit{Student} parameters.

We extend this paradigm to textual knowledge distillation by fixing both models and optimizing a natural-language representation, termed a \emph{Primer}, that enhances the \textit{Student} through context.
Let $\mathcal{V}$ denote the model vocabulary, $\mathcal{P}_{\mathrm{text}}\subseteq\mathcal{V}^{*}$ the candidate text space, and $P\oplus x$ the composition of a \emph{Primer} $P$ with an input $x$.
Textual knowledge distillation is formulated as
\begin{equation}
P^{*}
=
\arg\min_{P\in\mathcal{P}_{\mathrm{text}}}
\mathbb{E}_{x\sim\mathcal{D}}
\left[
\mathcal{F}_{\mathrm{KD}}
\left(
M_{\theta_S}(P\oplus x),
\mathcal{K}_{M_{\theta_T}}(x)
\right)
\right],
\qquad
\theta_T,\theta_S\ \text{fixed}.
\label{eq:textual-kd}
\end{equation}
Unlike Eq.~(\ref{eq:parametric-kd}), Eq.~(\ref{eq:textual-kd}) optimizes discrete text $P$ rather than real-valued parameters $\theta_S$, making the distilled knowledge auditable and reusable across models.
Because $\mathcal{P}_{\mathrm{text}}$ is discrete, this objective is typically approximated through search or LLM-based iteration.

For \textit{Primer} synthesis in UTT, the optimization data in
Eq.~(\ref{eq:textual-kd}) are instantiated as the distillation training\footnote{Although \textit{training} mostly means parameter updating in the conventional machine learning context, in this work, we use the term to mean the same as \textit{learning}, which is in the form of text (\textit{i.e.}, the \textit{Primer}), not weights.}
set $\mathcal{D}_{\mathrm{dist}}^{\mathrm{tr}}$ obtained through
knowledge-gap construction.
The criterion $\mathcal{F}_{\mathrm{KD}}$ denotes task-specific evaluation feedback, which may be numerical or textual.
The \textit{Teacher} signal $\mathcal{K}_{M_{\theta_T}}(x)$ combines evaluative and generative information.
The next section describes how the \emph{Primer} is generated through multi-role interaction.

\subsection{The UTT Framework}
\label{sec:framework}
\begin{figure*}[t]
\centering
\includegraphics[width=1\textwidth]{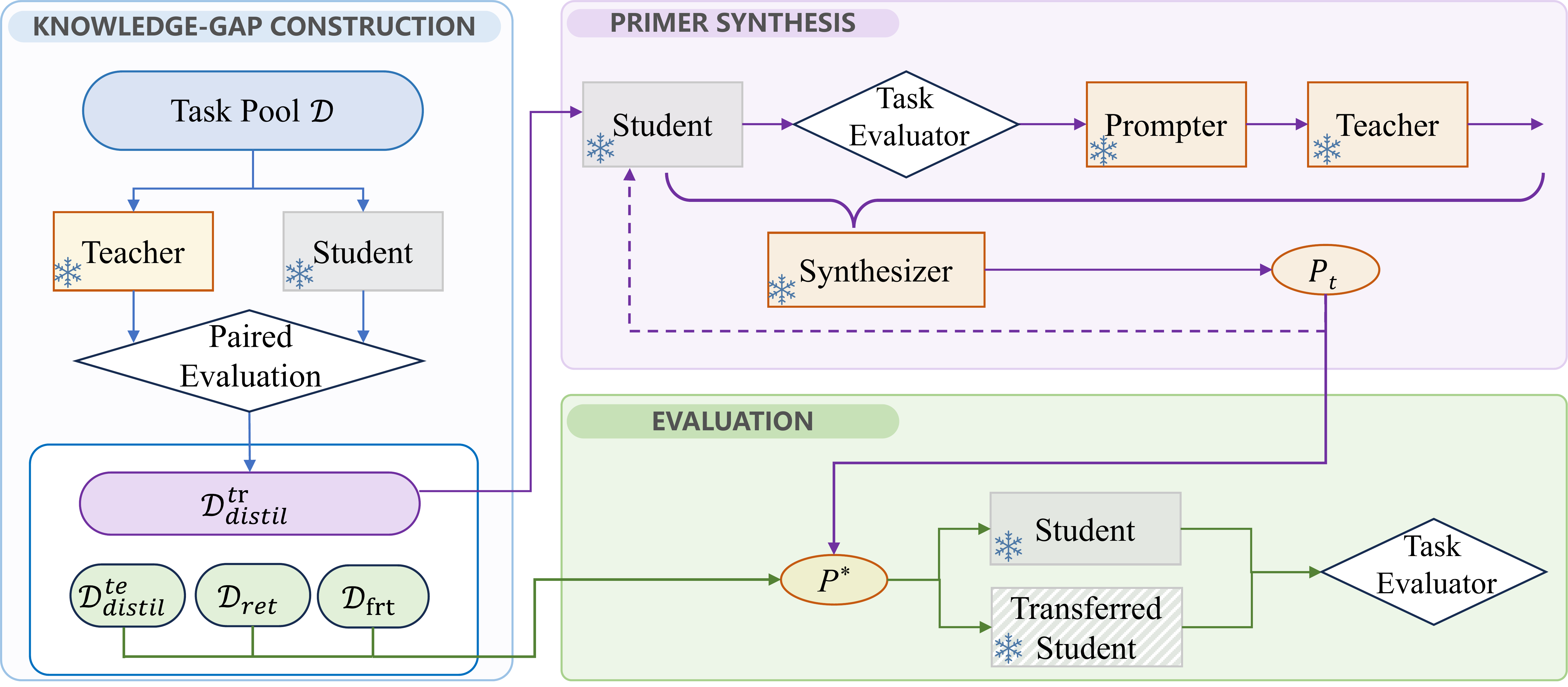}
\caption{\textbf{Overview of Universal Textual Teaching (UTT).}
\textbf{(1) Knowledge-gap construction.}
UTT first evaluates the knowledge gap between the \textit{Teacher} and \textit{Student}
and accordingly partitions $\mathcal{D}$ into subsets (Section~\ref{sec:gap_construction}).
\textbf{(2) Primer synthesis.}
It then synthesizes
a textual \textit{Primer} through iterative interactions among multiple
LLMs
(Section~\ref{sec:primer_synthesis}).
\textbf{(3) Evaluation.}
Finally, the resulting \textit{Primer} is applied to the source \textit{Student}
and transferred to other \textit{Students}, improving both the source
\textit{Student} and the transfer \textit{Students}.
All model parameters remain frozen throughout the process.
In the diagram, rectangles denote models, with the same color indicating
the same model serving different roles; rounded rectangles denote data
subsets, and diamonds denote evaluation procedures.}
\label{fig:UTT-framework}
\vspace{-2mm}
\end{figure*}

As shown in Figure~\ref{fig:UTT-framework}, UTT first partitions the
dataset through knowledge-gap construction and then synthesizes a
\textit{Primer} through multi-role LLM interaction. We detail both components below.

\subsubsection{Knowledge-Gap Construction}
\label{sec:gap_construction}

UTT first identifies the knowledge to transfer through paired evaluations.
Using all \textit{Teacher} generations would introduce redundant supervision, while a failing \textit{Teacher} cannot provide a reliable demonstration.
We therefore construct the knowledge-gap dataset from their observed performance.

Let $y_{m,x}^{(k)}$ denote the $k$-th generation of model $m\in\{T,S\}$ on sample $x$.
Let $\mathcal{F}_{\mathrm{KD}}^{\mathrm{corr}}(x,y)\in\{0,1\}$ denote the correctness field returned by $\mathcal{F}_{\mathrm{KD}}$, and define the empirical accuracy on $x$ as
\begin{equation}
q_m(x)
=
\frac{1}{K}
\sum_{k=1}^{K}
\mathcal{F}_{\mathrm{KD}}^{\mathrm{corr}}
\left(
x,y_{m,x}^{(k)}
\right),
\qquad m\in\{T,S\}.
\label{eq:empirical-accuracy}
\end{equation}

Based on $q_T(x)$ and $q_S(x)$, we partition $\mathcal{D}$ into three disjoint subsets: the \textit{distillation}, \textit{retention}, and \textit{frontier} subsets, denoted by $\mathcal{D}_{\mathrm{dist}}$, $\mathcal{D}_{\mathrm{ret}}$, and $\mathcal{D}_{\mathrm{frt}}$, respectively:
\begin{equation}
\begin{aligned}
\mathcal{D}
&=
\mathcal{D}_{\mathrm{dist}}
\mathbin{\dot{\cup}}
\mathcal{D}_{\mathrm{ret}}
\mathbin{\dot{\cup}}
\mathcal{D}_{\mathrm{frt}},\\
\mathcal{D}_{\mathrm{dist}}
&=
\left\{
x\in\mathcal{D}
\,\middle|\,
q_T(x)>q_S(x)
\right\},\\
\mathcal{D}_{\mathrm{ret}}
&=
\left\{
x\in\mathcal{D}
\,\middle|\,
q_T(x)\leq q_S(x),\ q_S(x)>0
\right\},\\
\mathcal{D}_{\mathrm{frt}}
&=
\left\{
x\in\mathcal{D}
\,\middle|\,
q_T(x)=q_S(x)=0
\right\}.
\end{aligned}
\label{eq:dataset-partition}
\end{equation}

The distillation subset $\mathcal{D}_{\mathrm{dist}}$ contains samples where the \textit{Teacher} outperforms the \textit{Student} for knowledge transfer, while the retention subset $\mathcal{D}_{\mathrm{ret}}$ and the frontier subset $\mathcal{D}_{\mathrm{frt}}$ evaluate capability preservation and potential improvement beyond the \textit{Teacher}.
$\mathcal{D}_{\mathrm{dist}}$ is further split into a distillation training set $\mathcal{D}_{\mathrm{dist}}^{\mathrm{tr}}$ and test set $\mathcal{D}_{\mathrm{dist}}^{\mathrm{te}}$.
Only the training set is used for \textit{Primer} synthesis, while all other subsets are reserved for final evaluation; namely, the training and testing sets do \textit{not} overlap, following the standard machine learning setting.

\subsubsection{Closed-Loop Primer Synthesis via Multi-Role Interaction}
\label{sec:primer_synthesis}
UTT processes the distillation training set in batches and progressively synthesizes global knowledge into the \textit{Primer}.
In each batch, the \textit{Student} role exposes residual knowledge gaps, the \textit{Prompter} role formulates teaching instructions, the \textit{Teacher} role provides targeted demonstrations, and the \textit{Synthesizer} role consolidates teaching records into a candidate \textit{Primer}.
Each role uses a task-agnostic system prompt and receives task-specific inputs at runtime (Appendix~\ref{app:role-prompts}).
All model parameters remain \textit{frozen} throughout the process.

Formally, let $\{\mathcal{B}_1,\ldots,\mathcal{B}_T\}$ denote the ordered batches of $\mathcal{D}_{\mathrm{dist}}^{\mathrm{tr}}$, and initialize the \textit{Primer} as $P_0$.
\paragraph{Multi-role teaching interaction.}
For each \(x_i\in\mathcal{B}_t\), the \textit{Student} first generates a trial
\(z_{t,i}\) using the current Primer \(P_{t-1}\), and the task evaluator
returns feedback \(e_{t,i}\). If the trial is correct, UTT proceeds
directly to the next problem.
Otherwise, the \textit{Prompter} \(M_{\theta_P}\) uses \(x_i\) and
\(e_{t,i}\) to formulate a teaching instruction \(a_{t,i}\), and the
\textit{Teacher} uses \(x_i\) and \(a_{t,i}\) to generate a targeted
demonstration \(d_{t,i}\):
\begin{equation}
\begin{aligned}
z_{t,i}&=M_{\theta_S}(P_{t-1}\oplus x_i),\quad
e_{t,i}=\mathcal{F}_{\mathrm{KD}}(x_i,z_{t,i}),\\
a_{t,i}&=M_{\theta_P}(x_i,e_{t,i}),\quad
d_{t,i}=M_{\theta_T}(x_i,a_{t,i}).
\end{aligned}
\label{eq:teaching-interaction}
\end{equation}
The interaction for an incorrect trial forms a teaching record
\(r_{t,i}\); to avoid introducing unreliable knowledge, only records whose
\textit{Teacher} demonstrations pass evaluation are retained in \(\mathcal{R}_t\):
\begin{equation}
\begin{aligned}
r_{t,i}
&=
\left(
x_i,z_{t,i},e_{t,i},a_{t,i},d_{t,i}
\right),\\
\mathcal{R}_t
&=
\left\{
r_{t,i}
\mid x_i\in\mathcal{B}_t,\,
\mathcal{F}_{\mathrm{KD}}^{\mathrm{corr}}(x_i,z_{t,i})=0,\,
\mathcal{F}_{\mathrm{KD}}^{\mathrm{corr}}(x_i,d_{t,i})=1
\right\}.
\end{aligned}
\label{eq:valid-teaching-records}
\end{equation}

\paragraph{\textit{Primer} synthesis and validation.}
The Synthesizer extracts textual knowledge from the $\mathcal{R}_t$ and integrates it with the current \textit{Primer} to produce a candidate $\widetilde{P}_t$:
\begin{equation}
\widetilde{P}_t
=
M_{\theta_{\mathrm{Syn}}}
\!\left(P_{t-1},\mathcal{R}_t\right).
\label{eq:primer-synthesis}
\end{equation}

To prevent accumulated knowledge from degrading \textit{Student} performance, UTT compares the current and candidate \textit{Primer} on the same batch.
Let $N_{\mathrm{corr}}(P;\mathcal{B}_t)$ denote the number of correct \textit{Student} outputs on $\mathcal{B}_t$ when using \textit{Primer} $P$:
\begin{equation}
N_{\mathrm{corr}}(P;\mathcal{B}_t)
=
\sum_{x_i\in\mathcal{B}_t}
\mathcal{F}_{\mathrm{KD}}^{\mathrm{corr}}
\left(
x_i,
M_{\theta_S}(P\oplus x_i)
\right).
\label{eq:primer-correct-count}
\end{equation}

The candidate passes the validation gate only if it performs no worse than the current \textit{Primer}:
\begin{equation}
P_t
=
\begin{cases}
\widetilde{P}_t,
&
N_{\mathrm{corr}}(\widetilde{P}_t;\mathcal{B}_t)
\geq
N_{\mathrm{corr}}(P_{t-1};\mathcal{B}_t),\\[1mm]
P_{t-1},
&
\text{otherwise}.
\end{cases}
\label{eq:primer-gate}
\end{equation}

After all batches are processed, the final global \textit{Primer} is obtained as $P^{*}$. During deployment, the \textit{Primer} can be directly provided to the \textit{Student} model without updating model parameters. Experiments further validate that this language-based knowledge representation \textit{Primer} not only enhances the source \textit{Student} but also transfers to improve other \textit{Student} models. 
Appendix~\ref{app:primer-examples} presents complete \textit{Primers} for Olympiad-level mathematics, GPU kernel generation, and visual geometry.

\section{Experimental Results}

\subsection{Experimental Settings}
\label{sec:exp-settings}
\textbf{Tasks and evaluation.}
We evaluate UTT on two representative LLM tasks, code generation and mathematical reasoning, using the challenging KernelBench \citep{kernelbench} and Omni-MATH-2 \citep{omni-math} benchmarks, respectively.
KernelBench is an open-ended benchmark without a unique reference implementation: it requires models to generate correct and efficient GPU kernels and evaluates their functionality and efficiency through execution, whereas Omni-MATH-2 consists of Olympiad-level mathematics problems with ground-truth answers.

Accuracy is the primary metric for both tasks. For KernelBench, we additionally report Fast$_1$@5, the fraction of problems with at least one of five samples that is correct and faster than the PyTorch reference.
We denote Fast$_1$@5 as Fast$_1$ hereafter.

\textbf{Comparison methods.}
We compare UTT with prompt engineering methods and parameter-updating KD methods.
Prompting baselines include one-shot, few-shot, teacher summary prompts, and automatic optimization methods such as APE \citep{ape}, MIPROv2 \citep{mipro}, and GEPA \citep{gepa}, while KD baselines include SeqKD \citep{seqkd}, Fine-tune-CoT \citep{Fine-tune-CoT}, RSR \citep{rsr}, and LUFFY \citep{luffy}.

\textbf{Roles and model configurations.}
In each configuration, the target student model serves as the \textit{Student}, while the source teacher model serves as the \textit{Teacher}, \textit{Prompter}, and \textit{Synthesizer}.
We use DeepSeek V4 Pro (Pro) and DeepSeek V4 Flash (Flash) \citep{deepseekv4}, Qwen3.6-27B (Qwen) \citep{qwen36_27b}, and Claude Opus 5 (Opus).
The three \textit{Teacher}-\textit{Student} configurations are Pro$\rightarrow$Flash, Pro$\rightarrow$Qwen, and Opus$\rightarrow$Flash.
They cover both within-family and cross-family knowledge transfer and include both open-source and proprietary models.

UTT uses a default batch size of 16, with evaluation budgets of 100 and 200 samples on KernelBench and Omni-MATH-2, respectively; detailed configurations are provided in Appendix~\ref{app:exp-settings}.

\subsection{Primer effectiveness and cross-model transferability}
\label{sec:exp-primer-main}

\begin{table}[t]
\centering
\caption{
Primer effectiveness and cross-model transferability.
For each task, each \textit{Teacher} and source \textit{Student} pair produces one
\textit{Primer}, transferring to the target \textit{Student}.
\textcircled{1}--\textcircled{6} denote different experimental groups.
Parentheses show gains over the corresponding \textit{Student}
baseline, and bold values indicate the best result.
All scores are computed over the full task pool $\mathcal{D}$.
}
\label{tab:primer-effectiveness-transfer}
\vspace{1mm}
{\begin{tabular*}{\linewidth}{@{\extracolsep{\fill}}lclcc@{}}
\toprule
\multicolumn{3}{c}{\textbf{Primer synthesis}}
& \multicolumn{2}{c}{\textbf{Target \textit{Student}}} \\
\cmidrule(lr){1-3}\cmidrule(lr){4-5}
\textit{Teacher}& \textit{Teacher} score & Source \textit{Student} & Flash & Qwen \\
\midrule
\multicolumn{5}{@{}l}{\textit{KernelBench Accuracy (\%)}} \\[1pt]
\multicolumn{3}{@{}l}{No \textit{Primer}}
& 9.4 & 9.2 \\

Pro & 19.6 & Flash
& \textbf{48.6 (+39.2)}\,\textcircled{1}
& 41.2 (+32.0)\,\textcircled{3} \\

Pro & 19.6 & Qwen
& 43.8 (+34.4)\,\textcircled{4}
& 50.0 (+40.8)\,\textcircled{2} \\

Opus & 36.0 & Flash
& 48.2 (+38.8)\,\textcircled{5}
& \textbf{51.0 (+41.8)}\,\textcircled{6} \\

\midrule
\multicolumn{5}{@{}l}{\textit{KernelBench Fast$_1$ (\%)}} \\[1pt]

\multicolumn{3}{@{}l}{No \textit{Primer}}
& 9 & 12 \\

Pro & 20 & Flash
& 35 (+26)\,\textcircled{1}
& \textbf{37 (+25)}\,\textcircled{3} \\

Pro & 20 & Qwen
& 30 (+21)\,\textcircled{4}
& 32 (+20)\,\textcircled{2} \\

Opus & 41 & Flash
& \textbf{39 (+30)}\,\textcircled{5}
& 22 (+10)\,\textcircled{6} \\

\midrule
\multicolumn{5}{@{}l}{\textit{Omni-MATH-2 Accuracy (\%)}} \\[1pt]

\multicolumn{3}{@{}l}{No \textit{Primer}}
& 27.6 & 34.3 \\

Pro & 51.4 & Flash
& \textbf{51.7 (+24.1)}\,\textcircled{1}
& 60.1 (+25.8)\,\textcircled{3} \\

Pro & 51.4 & Qwen
& 43.7 (+16.1)\,\textcircled{4}
& 62.4 (+28.1)\,\textcircled{2} \\

Opus & 92.5 & Flash
& 46.6 (+19.0)\,\textcircled{5}
& \textbf{65.8 (+31.5)}\,\textcircled{6} \\
\bottomrule
\end{tabular*}
\par}
\vspace{-2mm}
\end{table}

Table~\ref{tab:primer-effectiveness-transfer} summarizes the overall performance of the methods across different \textit{Teacher}-\textit{Student} configurations, evaluating the effectiveness and cross-model transferability of the \textit{Primer}.
We next analyze the results for each configuration and summarize the main findings.

\textbf{Basic Effectiveness and Student Replacement (\textcircled{1}, \textcircled{2}).}
We first evaluate Pro$\rightarrow$Flash, applying the
\textit{Primer} to its source \textit{Student}.
The \textit{Primer} yields substantial gains of $39.2$, $26$, and $24.1$
percentage points over the \textit{Student} baseline on KernelBench accuracy, Fast$_1$,
and Math accuracy, respectively, improving the corresponding scores from
$9.4\%$ to $48.6\%$, $9\%$ to $35\%$, and $27.6\%$ to $51.7\%$.
Keeping Pro as the Teacher, we further synthesize a new
\textit{Primer} for Qwen, which achieves $50.0\%$, $32\%$, and $62.4\%$
on the same metrics.
These results validate the effectiveness of UTT and demonstrate its applicability to \textit{Students} from different model families.

\textbf{Transfer across Students (\textcircled{3}, \textcircled{4}).}
To evaluate whether knowledge derived from the same \textit{Teacher} can be reused
across \textit{Students}, we exchange the two Pro-based \textit{Primers} between
Flash and Qwen without resynthesis.
The Flash-derived \textit{Primer} improves Qwen by $32.0$, $25$, and
$25.8$ percentage points in KernelBench accuracy, Fast$_1$, and Math accuracy,
respectively.
The Qwen-derived \textit{Primer} improves Flash by $34.4$, $21$, and
$16.1$ percentage points on the same metrics.
These results show that knowledge distilled through interactions with one
\textit{Student} can transfer to another \textit{Student} without resynthesizing the
\textit{Primer}.

\textbf{Teacher Replacement and General Transferability
(\textcircled{5}, \textcircled{6}).}
We then replace Pro with Opus and synthesize a new
\textit{Primer} for Flash.
It improves Flash by $38.8$, $30$, and $19.0$ percentage points on the
three metrics, demonstrating that UTT remains effective with different
\textit{Teachers}.
Finally, applying this \textit{Primer} unchanged to Qwen yields
$51.0\%$ KernelBench accuracy, $22\%$ Fast$_1$, and
$65.8\%$ Math accuracy.
Compared with the initial Pro$\rightarrow$Flash setting, both the \textit{Teacher} and
target \textit{Student} are changed, while Qwen does not participate in
\textit{Primer} synthesis.
This result further demonstrates the transferability and broad applicability
of UTT across diverse \textit{Teacher}-\textit{Student} configurations.

Together, these results support effective synthesis across
the evaluated model configurations without resynthesis or parameter updates. Although the main experiments focus on LLMs, we also preliminarily confirm transfer to multimodal LLMs in Appendix~\ref{app:multimodal}.

\subsection{Analysis: Generalization, Retention, and Frontier Extension}
\label{sec:exp-primer-analysis}
\begin{figure}[t]
\centering
\includegraphics[width=1\linewidth]{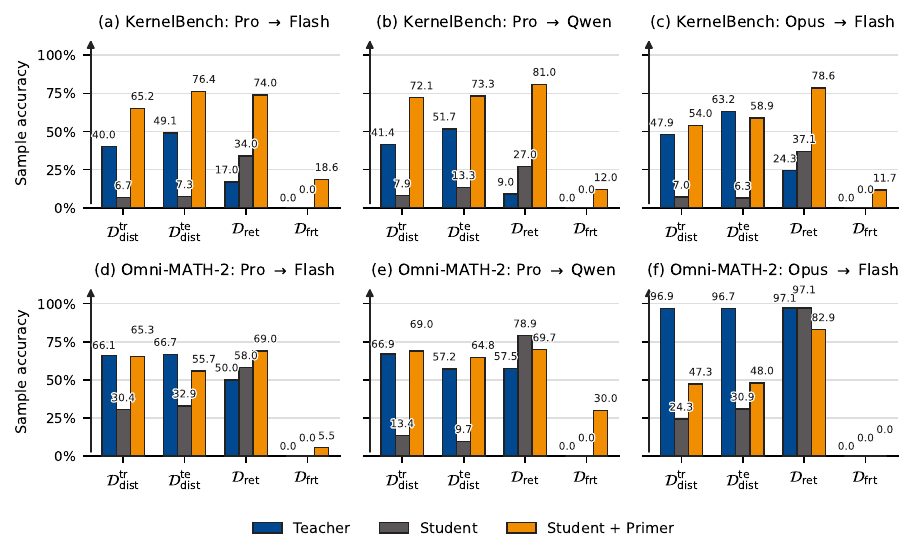}
\vspace{-4mm}
\caption{Split-wise sample accuracy for KernelBench and Omni-MATH-2.
Panels (a)--(c) show KernelBench results and panels (d)--(f)
show Omni-MATH-2 results.
From left to right, the three columns correspond to
Pro$\rightarrow$Flash, Pro$\rightarrow$Qwen, and
Opus$\rightarrow$Flash, respectively.
}
\label{fig:split-wise-analysis}
\vspace{-5mm}
\end{figure}
To analyze UTT under different knowledge-gap conditions in detail, Figure~\ref{fig:split-wise-analysis} reports sample accuracy for six task-model configurations across the four subsets constructed in Section~\ref{sec:gap_construction}, examining the effectiveness of \textit{Primer} text synthesis, the generalization of distilled knowledge, the preservation of existing capabilities, and the extension of capability frontiers, respectively.

On the \textbf{distillation} training set, Pro$\rightarrow$Flash accuracy on KernelBench increases from $6.7\%$ to $65.2\%$, indicating that the synthesized \textit{Primer} effectively improves \textit{Student} performance on knowledge-gap samples.
On the distillation test set, Pro$\rightarrow$Qwen accuracy on the math task increases from $9.7\%$ to $64.8\%$.
These results show that knowledge distilled into text can generalize to unseen problems, with benefits extending beyond the samples used for synthesis.

On the \textbf{retention} subset, the three KernelBench configurations show no decline in accuracy; instead, they improve from $34.0\%$ to $74.0\%$, $27.0\%$ to $81.0\%$, and $37.1\%$ to $78.6\%$, respectively.
On Omni-MATH-2, Pro$\rightarrow$Flash improves accuracy from $58.0\%$ to $69.0\%$, while Pro$\rightarrow$Qwen and Opus$\rightarrow$Flash decrease accuracy by $9.2$ and $14.2$ percentage points, respectively, yet still retain most of their baseline performance.
This indicates that the \textit{Primer} largely preserves the \textit{Student}'s existing capabilities while improving performance on knowledge-gap samples.

On the \textbf{frontier} subset, the primed \textit{Student} achieves nonzero accuracy in five of the six configurations.
The three KernelBench configurations achieve $11.7\%$ to $18.6\%$ accuracy, while Pro$\rightarrow$Flash and Pro$\rightarrow$Qwen achieve $5.3\%$ and $30.0\%$ on Omni-MATH-2, respectively.
This shows that the \textit{Primer} can help the \textit{Student} solve some problems that neither the \textit{Teacher} nor the \textit{Student} solved during baseline evaluation, extending the observed capability frontier.

Together, these fine-grained results further support the effectiveness of UTT in four respects: synthesis effectiveness, generalization to unseen problems, preservation of existing capabilities, and extension of capability frontiers.

\subsection{Comparative Experiments}
\label{sec:exp-comparison}
\begin{table}[!t]
\centering
\caption{
Comparison with prompt engineering methods.
``\textit{Teacher} summary'' means asking the \textit{Teacher} to summarize the training set into a
general prompt.
}
\label{tab:prompt-engineering-comparison}

{\footnotesize
\setlength{\tabcolsep}{3.2pt}
\renewcommand{\arraystretch}{0.98}
\setlength{\aboverulesep}{0.35ex}
\setlength{\belowrulesep}{0.45ex}
\vspace{1mm}
\begin{tabular*}{\linewidth}{@{\extracolsep{\fill}}lcccccc@{}}
\toprule
\multirow{3}{*}{\textbf{Method}}
& \multicolumn{2}{c}{\textbf{Omni-MATH-2}}
& \multicolumn{4}{c}{\textbf{KernelBench}} \\
\cmidrule(lr){2-3}\cmidrule(lr){4-7}
& \textit{Flash}
& \textit{Qwen}
& \multicolumn{2}{c}{\textit{Flash}}
& \multicolumn{2}{c}{\textit{Qwen}} \\[1pt]
& Acc.
& Acc.
& Acc.
& Fast$_1$
& Acc.
& Fast$_1$ \\
\midrule

\textit{Student}
& 27.6
& 34.3
& 9.4
& 9
& 9.2
& 12 \\

\midrule
\multicolumn{7}{@{}l}{\textit{Prompting Techniques (PT)}} \\[1pt]

\quad One-shot
& 35.0
& 57.7
& 39.2
& 23
& 45.8
& 31 \\

\quad Few-shot
& 35.4
& 54.8
& 40.6
& 25
& 46.6
& 30 \\

\quad \textit{Teacher} summary
& 36.7
& 32.8
& 40.0
& 30
& 10.2
& 11 \\

\midrule
\multicolumn{7}{@{}l}{\textit{Automatic Prompt Optimization (APO)}} \\[1pt]

\quad APE \citep{ape}
& 31.0
& 39.3
& 11.0
& 8
& 10.4
& 13 \\

\quad MIPROv2 (\textit{instruction only})
& 28.7
& 29.8
& 9.8
& 6
& 11.8
& 9 \\

\quad MIPROv2 \citep{mipro}
& 33.5
& 25.8
& 45.0
& 21
& 44.6
& 30 \\

\quad GEPA \citep{gepa}
& 41.1
& 61.9
& 34.8
& 24
& 32.6
& 23 \\

\midrule

\textbf{UTT (Ours)}
& \textbf{51.7}
& \textbf{62.4}
& \textbf{48.6}
& \textbf{35}
& \textbf{50.0}
& \textbf{32} \\

\bottomrule
\end{tabular*}
}
\end{table}

\textbf{Comparison with prompt engineering methods.}
Table~\ref{tab:prompt-engineering-comparison} compares UTT with several prompt engineering methods. UTT performs best across all six combinations, with consistent gains across tasks and target models.
Among the prompting baselines, GEPA is stronger on Omni-MATH-2, while MIPROv2 performs better on KernelBench.
\textit{Teacher} summary improves Flash but is inconsistent on Qwen, suggesting that directly summarized \textit{Teacher} knowledge may not transfer well across \textit{Students} and \textit{Teacher}-\textit{Student} interaction is important for constructing an effective \textit{Primer}.
Appendix~\ref{app:long-token-test} further examines this trend under larger output-token budgets for math.

\begin{table}[!t]
\caption{
Comparison with KD methods under the
Pro$\rightarrow$Qwen setting.
}
\label{tab:kd-comparison}

\begingroup
\footnotesize
\setlength{\tabcolsep}{4pt}
\renewcommand{\arraystretch}{0.96}
\setlength{\aboverulesep}{0.35ex}
\setlength{\belowrulesep}{0.45ex}

\noindent
\vspace{1mm}
\begin{tabular*}{\linewidth}{@{\extracolsep{\fill}}lccc@{}}
\toprule
\multirow{2}{*}{\textbf{Method}}
& \textbf{Omni-MATH-2}
& \multicolumn{2}{c}{\textbf{KernelBench}} \\
\cmidrule(lr){2-2}\cmidrule(lr){3-4}
& Acc. & Acc. & Fast$_1$ \\
\midrule

\textit{Student}
& 34.3 & 9.2 & 12 \\

\midrule

SeqKD \citep{seqkd}
& 53.5 & 18.8 & 18 \\

Fine-tune-CoT \citep{Fine-tune-CoT}
& 39.5 & 41.8 & 28 \\

RSR \citep{rsr}
& 51.5 & 22.6 & 19 \\

LUFFY \citep{luffy}
& 55.2 & 11.4 & 13 \\

\midrule

\textbf{UTT (Ours)}
& \textbf{62.4}
& \textbf{50.0}
& \textbf{32} \\

\bottomrule
\end{tabular*}

\endgroup
\vspace{-3mm}
\end{table}

\textbf{Comparison with KD methods.}
Table~\ref{tab:kd-comparison} compares UTT with parameter-updating KD methods.
With \textit{Student} parameters fixed, UTT exceeds the strongest baseline by $7.2$, $8.2$, and $4$ percentage points on Omni-MATH-2 accuracy, KernelBench accuracy, and Fast$_1$, respectively.
These results indicate that explicit textual knowledge can yield stronger and more consistent gains across tasks.
Appendix~\ref{app:computational-cost} reports the resources used to achieve these results.

We have more results, but due to page limits, ablations of knowledge-gap selection, the \textit{Prompter}, and validation gating, along with analyses of batch size, quantization, and thinking mode, are deferred to Appendix~\ref{app:ablations}. Potential limitations and future work are discussed in Appendix~\ref{app:limitation_future_work}.

\section{Conclusion}
This paper introduces \textit{Universal Textual Teaching} (UTT), a parameter-update-free knowledge distillation method for large language models. UTT first identifies \textit{Teacher}-\textit{Student} knowledge gaps and then synthesizes validated teaching records into a textual \textit{Primer} through multi-role LLM interactions.
On the challenging math and code generation tasks (Omni-MATH-2 and KernelBench), UTT substantially improves the \textit{Student}'s performance across various \textit{Teacher-Student} settings. UTT also outperforms representative prompt engineering and KD methods.
Compared with knowledge implicitly encoded in model parameters, the \textit{Primer} is interpretable and not tied to particular model weights, enabling direct transfer to other models.
This demonstrates that natural language can serve as a deployable knowledge carrier across models, providing a new parameter-update-free perspective on knowledge distillation for large language models.


\clearpage
\bibliography{references}
\bibliographystyle{references}

\clearpage
\appendix

\section{Experimental Settings}
\label{app:exp-settings}

\subsection{Datasets and Evaluation}

To evaluate knowledge transfer on challenging tasks with capability gaps between the \textit{Teacher} and \textit{Student}, we use KernelBench \citep{kernelbench} and Omni-MATH-2 \citep{omni-math} as our primary benchmarks.
For KernelBench, we use all 100 problems from Level 1.
For Omni-MATH-2, we select a fixed set of 200 problems from difficulty levels 5--10, balanced across categories such as geometry, algebra, and number theory.
To reduce the effect of stochasticity from API services, each model independently generates five responses per problem, resulting in initial pools of 500 KernelBench samples and 1,000 Omni-MATH-2 samples.

Both tasks follow the official evaluation protocols of their respective benchmarks.
For Omni-MATH-2, we report final-answer accuracy.
KernelBench evaluates generated kernels through execution-based functional correctness checks and measures their runtime efficiency relative to the PyTorch reference implementations.
Because KernelBench runtimes depend on the underlying hardware, all generated kernels and PyTorch reference implementations are evaluated on an NVIDIA RTX A6000.
In addition to sample-level functional accuracy, we report Fast$_1$@5, defined as the fraction of problems for which at least one of five generated samples is functionally correct and faster than the PyTorch reference implementation.
For comparison with the \textit{Student} baseline used during knowledge-gap construction, the primed model regenerates the full task pool under the same configuration with five samples per problem.

\subsection{Models and Role Assignments}

We distinguish an underlying model from the functional role it performs
in the multi-role interaction.
For each \textit{Teacher}-\textit{Student} configuration, the target
\textit{Student} model instantiates the \textit{Student} role, whereas
the \textit{Teacher} model instantiates the \textit{Teacher}, \textit{Prompter}, and
\textit{Synthesizer} roles.
The roles use distinct prompts and runtime inputs but no role-specific
model configurations: all calls to the same underlying model use the
decoding settings reported below.
No model parameters are updated throughout UTT.

We consider three \textit{Teacher}-\textit{Student} configurations: DeepSeek V4 Pro $\rightarrow$ DeepSeek V4 Flash, DeepSeek V4 Pro $\rightarrow$ Qwen3.6-27B, and Claude Opus 5 $\rightarrow$ DeepSeek V4 Flash.
We abbreviate these models as Pro, Flash, Qwen, and Opus, respectively.

All models operate with thinking enabled.
DeepSeek uses the high thinking setting, Claude Opus 5 uses adaptive thinking, and Qwen uses its default thinking configuration.
The temperature is set to 0 for DeepSeek and Qwen, corresponding to greedy decoding, while no temperature is explicitly specified for Opus.
The maximum output length is set to 16,384 tokens for KernelBench, which is sufficient for complete GPU kernel generation.
For Omni-MATH-2, the maximum output length is set to 32,768 tokens.

\subsection{Role Prompt Templates}
\label{app:role-prompts}

The following canonical templates define the behavior and input-output
interface of the four roles in UTT.
Fields enclosed in braces are filled at runtime, and task-specific
requirements, including the required output format, are included in
\texttt{\{task\}}.
For clarity, each listing presents the system instruction together with
its runtime input fields; the implementation sends these two parts as
the system and user messages, respectively.
For batched synthesis, the case-level fields are repeated for each
teaching record in the batch.

\begin{UTTRolePromptBox}{\textit{Student} Role Prompt}
You are the Student. Solve the given task correctly and as well as you can.

The current Primer contains reusable knowledge, reasoning strategies, validation methods, and common pitfalls distilled from previous teaching cases. Use any relevant information from it, but do not treat it as a restriction.

Follow all requirements contained in the task. Output only the solution, with no additional commentary.

CURRENT PRIMER:
{current_primer}

TASK:
{task}
\end{UTTRolePromptBox}

\begin{UTTRolePromptBox}{\textit{Prompter} Role Prompt}
You are the Prompter. A Student is attempting a task, and a Teacher will prepare a worked solution to help the Student improve.

Based on the task and the evaluation feedback from the Student's most recent attempt, identify the primary knowledge or reasoning gap. Distinguish the underlying cause from its symptoms, and state what the Teacher's worked solution should emphasize.

Ground the instruction in the evaluation results and diagnostics. Focus on transferable principles, reasoning strategies, checks, or decision rules that would help with related tasks.

Do not solve the task yourself. Output only a concrete and actionable teaching instruction in concise sentences.

TASK:
{task}

EVALUATION FEEDBACK:
{feedback}
\end{UTTRolePromptBox}

\begin{UTTRolePromptBox}{\textit{Teacher} Role Prompt}
You are the Teacher. Produce a correct, complete, and high-quality worked solution to the given task.

Follow the teaching instruction and emphasize the knowledge, reasoning strategy, or validation steps it identifies. The solution should help the Student learn and should contain useful knowledge.

Follow all output-format requirements contained in the task. Output only the worked solution, with no additional commentary.

TASK:
{task}

TEACHING INSTRUCTION:
{instruction}
\end{UTTRolePromptBox}

\begin{UTTRolePromptBox}{\textit{Synthesizer} Role Prompt}
You are the Synthesizer. Maintain a single evolving Primer that captures transferable knowledge for solving a family of related tasks.

Use the task, teaching instruction, worked solution, and evaluation feedback to identify useful concepts, reasoning strategies, decision rules, validation methods, and common failure modes. Integrate these lessons into the current Primer.

Merge new knowledge into the appropriate existing sections rather than appending a history of edits. Generalize beyond the specific task while remaining concrete and actionable. Preserve useful existing knowledge, combine duplicates, and remove content only when it is redundant, incorrect, or too vague to be useful.

The Primer is a compact study guide, not a solution archive or changelog. Do not copy the complete task solution or retain unnecessary task-specific details. Keep the revised Primer within the specified word limit.

Output only the complete revised Primer, with no preamble or explanation of the changes.

TASK:
{task}

TEACHING INSTRUCTION:
{instruction}

WORKED SOLUTION:
{demo}

EVALUATION FEEDBACK:
{feedback}

CURRENT PRIMER:
{current_primer}

WORD LIMIT:
{max_words}
\end{UTTRolePromptBox}

\subsection{Method and Baseline Configurations}

\textbf{UTT configuration.}
Based on paired \textit{Teacher} and \textit{Student} performance,
UTT constructs distillation, retention, and frontier
subsets and divides the distillation subset into training and test
portions using a $7{:}3$ ratio.
The distillation-training portion is used for \textit{Primer} synthesis,
whereas the distillation-test portion is used for evaluation.
Table~\ref{tab:knowledge-gap-split-sizes} reports the number of generated
samples.

\begin{table}[!htbp]
\centering
\caption{
Numbers of generated samples associated with the knowledge-gap
portions across tasks and Teacher--\textit{Student} configurations.
}
\label{tab:knowledge-gap-split-sizes}

{\footnotesize
\setlength{\tabcolsep}{3.0pt}
\renewcommand{\arraystretch}{1.05}

\begin{tabular*}{\linewidth}{
@{\extracolsep{\fill}}
llrrrrr
@{}
}
\toprule
\textbf{Task}
& \textbf{Teacher$\rightarrow$\textit{Student}}
& $\mathcal{D}_{\mathrm{dist}}^{\mathrm{tr}}$
& $\mathcal{D}_{\mathrm{dist}}^{\mathrm{te}}$
& $\mathcal{D}_{\mathrm{ret}}$
& $\mathcal{D}_{\mathrm{frt}}$
& \textbf{Total} \\
\midrule

\multirow{3}{*}{KernelBench}
& Pro$\rightarrow$Flash
& 135 & 55 & 100 & 210
& \multirow{3}{*}{500} \\
& Pro$\rightarrow$Qwen
& 140 & 60 & 100 & 200
& \\
& Opus$\rightarrow$Flash
& 215 & 95 & 70 & 120
& \\

\midrule

\multirow{3}{*}{Omni-MATH-2}
& Pro$\rightarrow$Flash
& 490 & 210 & 100 & 200
& \multirow{3}{*}{1000} \\
& Pro$\rightarrow$Qwen
& 335 & 145 & 360 & 160
& \\
& Opus$\rightarrow$Flash
& 645 & 275 & 35 & 45
& \\

\bottomrule
\end{tabular*}
}
\end{table}

Closed-loop \textit{Primer} synthesis uses a default batch size of 16.
At inference time, the \textit{Primer} is prepended to each task input.
To prevent unbounded growth across synthesis rounds, the Synthesizer is instructed to respect a configurable soft word ceiling: 1,000 words by default.
Each accepted \textit{Primer} is carried forward to the next batch, and the final accepted version is used for that configuration.
The optimization budget is counted by sample-level evaluations performed during optimization: 100 evaluations for KernelBench and 200 for Omni-MATH-2.

\textbf{Prompting baselines.}
For KernelBench, the one-shot and few-shot baselines use the official benchmark prompt templates containing one and three demonstrations, respectively.
For Omni-MATH-2, we randomly select three examples with ground-truth answers from the source dataset outside the selected task pool.
\textit{Teacher} summary uses the \textit{Teacher} model to summarize the distillation training problems into a global prompt.

\textbf{APO baseline settings.}
APE \citep{ape}, MIPROv2 \citep{mipro}, and GEPA \citep{gepa} use their standard implementations, with Pro generating and refining candidate prompts and Flash or Qwen serving as the corresponding task model.
All three methods use the same distillation training pool, from which 20 problems for Omni-MATH-2 and 10 for KernelBench are held out as their validation sets.
Final prompts are selected according to validation performance.
MIPROv2 is evaluated in both instruction-only and demonstration-augmented settings; the latter uses two demonstrations whose responses are generated by the corresponding \textit{Student} and filtered by the task evaluator.
For each task, all APO baselines and UTT are compared under the same total optimization budget.

\textbf{KD baseline settings.}
Under Pro$\rightarrow$Qwen, the parameter-updating KD baselines comprise SeqKD \citep{seqkd}, Fine-tune-CoT \citep{Fine-tune-CoT}, RSR \citep{rsr}, and LUFFY \citep{luffy}, all using the same distillation training set as UTT.
All methods train LoRA adapters on a frozen Qwen3.6-27B backbone, with rank 16 and alpha 32, applied to 496 language-model projection layers while excluding the vision modules, embedding layers, and language-model head.

SeqKD, Fine-tune-CoT, and RSR optimize an autoregressive cross-entropy objective with AdamW, using a learning rate of $2\times10^{-4}$, LoRA dropout of 0.05, zero weight decay, and gradient-norm clipping at 1.0.
LUFFY uses a grouped policy-gradient objective over on-policy \textit{Student} trajectories, with a learning rate of $1\times10^{-6}$ and dropout disabled.
The KD baselines perform 200 parameter updates on Omni-MATH-2 and 100 on KernelBench, with the final checkpoint used for evaluation.
We observed that the training losses of the KD baselines had largely stabilized by the end of their respective update budgets.

\section{Additional Experiments and Analysis}
\label{app:additional-experiments}
\subsection{Extension to Multimodal Reasoning}
\label{app:multimodal}

To preliminarily evaluate UTT in multimodal reasoning settings, we select 100 visual geometry problems from the Vision Dominant subset of MathVerse \citep{zhang2024mathverse} and independently sample two responses per problem.
We consider three Teacher--\textit{Student} configurations: Qwen3.8-Max$\rightarrow$Qwen3.6-27B, Qwen3.8-Max$\rightarrow$DeepSeek-V4.1-Flash, and Claude Opus 5$\rightarrow$Qwen3.6-27B.
Temperature is set to 0 for all models except Claude Opus 5, for which no temperature value is specified. The Qwen models are accessed through their official APIs, and the maximum output budget is set to 4,096 tokens.
The synthesis procedure follows the main experiments.

\begin{table}[t]
\centering
\caption{
Primer effectiveness and cross-model transferability on Vision Dominant plane geometry problems from MathVerse.
Each model is evaluated using two samples per problem.
\textit{Teacher} scores denote \textit{Teacher} accuracy.
Parentheses show percentage-point gains over the corresponding target \textit{Student} baseline, and bold values indicate the best result for each target \textit{Student}.
Max, Qwen, Flash 4.1, and Opus denote Qwen3.8-Max, Qwen3.6-27B, DeepSeek-V4.1-Flash, and Claude Opus 5, respectively.
}
\label{tab:multimodal-results}

{\footnotesize
\setlength{\tabcolsep}{4.5pt}
\renewcommand{\arraystretch}{1.0}

\begin{tabular*}{\linewidth}{
@{\extracolsep{\fill}}lclcc@{}
}
\toprule
\multicolumn{3}{c}{\textbf{Primer synthesis}}
& \multicolumn{2}{c}{\textbf{Target \textit{Student}}} \\
\cmidrule(lr){1-3}
\cmidrule(lr){4-5}
Teacher
& \textit{Teacher} score
& Source \textit{Student}
& Flash 4.1
& Qwen \\
\midrule

\multicolumn{3}{@{}l}{No \textit{Primer}}
& 54.0
& 41.5 \\

Max
& 94.0
& Qwen
& \textbf{59.0 (+5.0)}
& 76.0 (+34.5) \\

Max
& 94.0
& Flash 4.1
& 56.0 (+2.0)
& 72.0 (+30.5) \\

Opus
& 93.5
& Qwen
& 57.5 (+3.5)
& \textbf{80.5 (+39.0)} \\

\bottomrule
\end{tabular*}
\par
}
\end{table}

Table~\ref{tab:multimodal-results} shows that the \textit{Primer} yields positive gains for every source and transfer \textit{Student}.
Under Qwen3.8-Max$\rightarrow$Qwen3.6-27B, the source \textit{Student} improves from $41.5\%$ to $76.0\%$, while transferring the same \textit{Primer} to DeepSeek-V4.1-Flash improves its accuracy from $54.0\%$ to $59.0\%$.
When DeepSeek-V4.1-Flash serves as the source \textit{Student}, its accuracy increases from $54.0\%$ to $56.0\%$, whereas transferring the resulting \textit{Primer} to Qwen3.6-27B substantially improves accuracy from $41.5\%$ to $72.0\%$.
After replacing the \textit{Teacher} with Claude Opus 5, Qwen3.6-27B improves from $41.5\%$ to $80.5\%$, while applying the same \textit{Primer} to DeepSeek-V4.1-Flash raises its accuracy from $54.0\%$ to $57.5\%$.

Overall, \textit{Primers} synthesized by different Teachers substantially improve Qwen3.6-27B, and all cross-Student transfers retain positive gains.
These results indicate that UTT extends to multimodal reasoning tasks requiring joint understanding of images and text, and that its textual knowledge can be reused across models.
The different gain magnitudes on Qwen3.6-27B and
DeepSeek-V4.1-Flash indicate that transfer effectiveness varies with the
target \textit{Student}.

\subsection{Primer Effectiveness under Larger Token Budgets}
\label{app:long-token-test}
\begin{figure}[t]
\centering
\includegraphics[width=\linewidth]{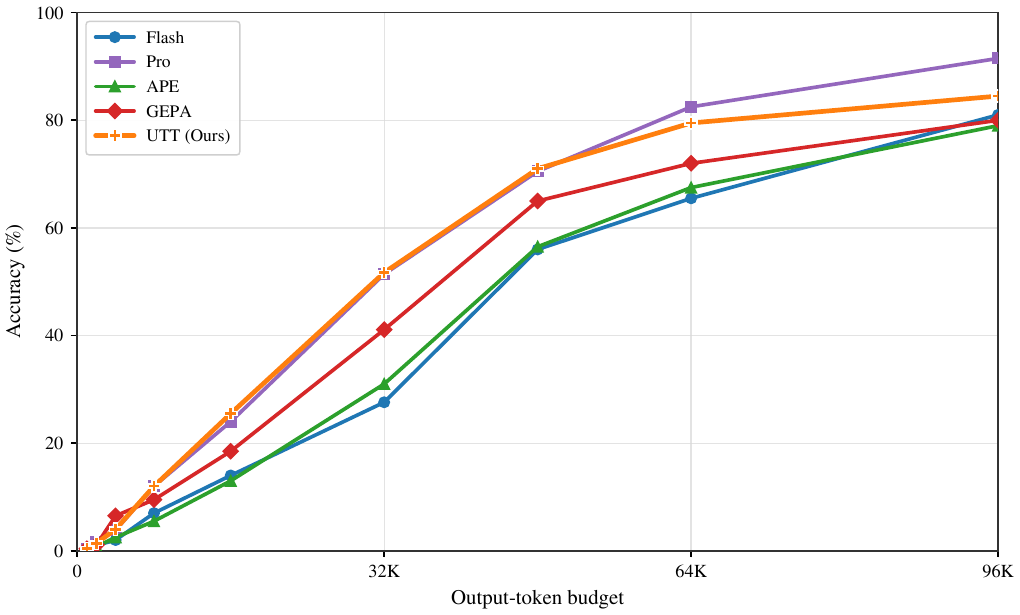}
\caption{
Effect of output-token budget on Omni-MATH-2 accuracy.
Curves compare the unprimed Flash \textit{Student}, the Pro Teacher, APE, GEPA, and UTT.
UTT uses the same \textit{Primer} synthesized under the 32k-token budget at every evaluated budget, without resynthesis.
}
\label{fig:long-token-test}
\end{figure}

To examine how inference length affects performance, we evaluate the
unprimed \textit{Student}, prompt-optimization methods, UTT, and the
\textit{Teacher} under progressively larger maximum output-token budgets.
To isolate the effect of inference budget from that of \textit{Primer}
synthesis, UTT uses the same \textit{Primer} synthesized under the
32k-token setting at every evaluated budget, without resynthesis.

As shown in Figure~\ref{fig:long-token-test}, accuracy generally
increases with the output-token budget, while the incremental gains
diminish at larger budgets, indicating that performance gradually
approaches saturation.
UTT consistently outperforms the unprimed \textit{Student} across all
evaluated budgets, with a more pronounced advantage under smaller
budgets.
As both curves approach saturation, UTT retains its advantage,
suggesting that the \textit{Primer} is particularly beneficial when
the inference budget is constrained and provides a higher effective
performance ceiling within the evaluated range.

As shown in Figure~\ref{fig:long-token-test}, errors caused by truncation gradually decrease as the output-token budget increases.
Notably, although the \textit{Primer} is synthesized with a 32k maximum output-token budget, it continues to improve the \textit{Student} at larger budgets, indicating that its knowledge and solution strategies are not tied to the specific output-token budget used during synthesis.
At the largest tested budget, APO approaches and even slightly underperforms the baseline, whereas UTT, although not reaching \textit{Teacher} performance, remains between the baseline and the \textit{Teacher}.
This suggests that explicit knowledge guidance derived from \textit{Teacher}-\textit{Student} interaction can narrow this gap more effectively than prompt optimization that primarily relies on iterative search guided by evaluation signals.

\subsection{Resource Comparison between KD Training and UTT Primer Synthesis}
\label{app:computational-cost}

\begin{table}[t]
\centering
\caption{
Stage-specific resource usage for the Omni-MATH-2 results in
Table~\ref{tab:kd-comparison} under the Pro$\rightarrow$Qwen setting.
We treat the task data as given and focus on the method-specific
optimization stage: KD reports parameter training only, whereas UTT
reports the time, model calls, and tokens used for API-based
\textit{Primer} synthesis.
Because the two method families consume different resource types,
these measurements characterize their respective resource profiles
rather than constituting a strictly equivalent end-to-end cost comparison.
\textit{Teacher} side aggregates the Teacher, Prompter, and Synthesizer.
}
\label{tab:computational-cost}

{\footnotesize
\setlength{\tabcolsep}{3.5pt}
\renewcommand{\arraystretch}{1.05}

\begin{tabular*}{\linewidth}{
@{\extracolsep{\fill}}
llcrrr
@{}
}
\toprule
\textbf{Method / Role}
& \textbf{Compute}
& \shortstack{\textbf{Time}\\\textbf{(h)}}
& \shortstack{\textbf{Model}\\\textbf{calls}}
& \shortstack{\textbf{Input}\\\textbf{tokens}}
& \shortstack{\textbf{Output}\\\textbf{tokens}} \\
\midrule

\multicolumn{6}{@{}l}{\textit{Parameter-updating KD}} \\[1pt]

\quad SeqKD
& 4$\times$A6000
& 1.2
& -- & -- & -- \\

\quad Fine-tune-CoT
& 4$\times$A6000
& 3.5
& -- & -- & -- \\

\quad RSR
& 4$\times$A6000
& 7.5
& -- & -- & -- \\

\quad LUFFY
& 8$\times$A6000
& 22.0
& -- & -- & -- \\

\midrule

\multicolumn{6}{@{}l}{\textit{UTT}} \\[1pt]

\quad \textbf{Total}
& Official APIs
& 1.6
& 281
& 677,591
& 5,568,109 \\

\quad \textit{Teacher} side
& --
& --
& 70
& 194,043
& 279,483 \\

\quad \textit{Student} side
& --
& --
& 211
& 483,548
& 5,288,626 \\

\bottomrule
\end{tabular*}
\par
}
\end{table}

Assuming fixed and available task data, we compare the resources consumed
by the method-specific optimization stage required to obtain the trained
models or the \textit{Primer} evaluated in
Table~\ref{tab:kd-comparison}.
For parameter-updating KD, we report only the elapsed time of parameter
training. SeqKD, Fine-tune-CoT, and RSR use four RTX A6000 GPUs, while
LUFFY uses eight; these GPUs are used only to reproduce the KD baselines.
UTT performs no parameter training and uses no local GPUs; its reported
time, model calls, and token counts correspond to \textit{Primer}
synthesis through official APIs.

As shown in Table~\ref{tab:computational-cost}, parameter training for
the three supervised KD baselines requires $1.2$--$7.5$ hours, while
the reinforcement-learning-based LUFFY requires $22.0$ hours on eight
GPUs.
UTT completes \textit{Primer} synthesis in $1.6$ hours through 281 API
model calls, approximately 0.68 million input tokens, and 5.57 million
output tokens.
Together with Table~\ref{tab:kd-comparison}, these results show that UTT
outperforms all KD baselines without parameter updates. However, because
the two method families primarily consume GPU training compute and API
inference tokens, respectively, we do not interpret their resource
measurements as strictly equivalent costs.

\section{Ablation Studies}
\label{app:ablations}

\subsection{Effects of Core Components and Batch Size}
\label{app:core-ablations}

\textbf{Experimental setup.}
We examine knowledge-gap selection, the \textit{Prompter}, validation gating, and
batch size on Omni-MATH-2 under the Pro$\rightarrow$Flash configuration.
The full method uses 98 knowledge-gap training problems, retains the
Prompter and validation gate, and sets the default batch size to 16.
All variants use 98 synthesis problems.
For final evaluation, each ablation variant independently generates two responses per problem, with all other generation settings following the main experiments.
All variants are evaluated on the four subsets from the original
partition; the distillation-test, retention, and frontier subsets are
excluded from \textit{Primer} synthesis in every variant.

\textbf{Component ablations.}
The random-selection variant samples 98 problems from a larger candidate
pool drawn from the original dataset, excluding all problems in the fixed
distillation-test, retention, and frontier subsets.
This variant preserves the number of synthesis problems but removes
selection based on the Teacher--\textit{Student} knowledge gap.
The variant without the Prompter omits the teaching instruction and asks
the \textit{Teacher} to demonstrate directly from the problem, \textit{Student} attempt,
and evaluation feedback.
The variant without validation gating still evaluates and records both
\textit{Primer} versions on the same batch, but accepts candidate updates
regardless of the comparison.
Correctness filtering of \textit{Teacher} demonstrations remains unchanged across
these variants.

\textbf{Batch size.}
We additionally evaluate batch sizes of 8 and 32 while keeping the
training problems, their order, and all other settings fixed.
These variants examine the number of teaching records consolidated per
update and the resulting update frequency.

\begin{table}[t]
\centering
\caption{
Core-component and batch-size ablations under Pro$\rightarrow$Flash on
Omni-MATH-2.
Bold values indicate the best result for each subset.
}
\label{tab:core-ablations}

{\footnotesize
\setlength{\tabcolsep}{3pt}
\renewcommand{\arraystretch}{1.05}

\begin{tabular*}{\linewidth}{
@{\extracolsep{\fill}}lccccc@{}
}
\toprule
\textbf{Variant}
& \textbf{Batch}
& $\mathcal{D}_{\mathrm{dist}}^{\mathrm{tr}}$
& $\mathcal{D}_{\mathrm{dist}}^{\mathrm{te}}$
& $\mathcal{D}_{\mathrm{ret}}$
& $\mathcal{D}_{\mathrm{frt}}$ \\
\midrule

\textit{Student} without \textit{Primer}
& -- & 33.7 & 33.3 & 60.0 & 0.0 \\

\midrule

Full method
& 16 & 65.3 & \textbf{56.0} & 72.5 & \textbf{5.0} \\

Random selection
& 16 & 55.6 & 50.0 & 72.5 & 3.8 \\

Without \textit{Prompter}
& 16 & \textbf{72.4} & 48.8 & 70.0 & \textbf{5.0} \\

Without validation gate
& 16 & 59.2 & 39.3 & \textbf{80.0} & 3.8 \\

\midrule

Batch size 8
& 8 & 53.6 & 41.7 & 67.5 & 2.5 \\

Batch size 32
& 32 & 59.2 & 45.2 & 72.5 & \textbf{5.0} \\

\bottomrule
\end{tabular*}
}
\end{table}

\textbf{Overall effectiveness.}
The full method improves over the \textit{Student} without a
\textit{Primer} across all four knowledge-gap subsets.
Accuracy increases from $33.7\%$, $33.3\%$, $60.0\%$, and $0.0\%$ to
$65.3\%$, $56.0\%$, $72.5\%$, and $5.0\%$ on the distillation-training,
distillation-test, retention, and frontier subsets, respectively.

\textbf{Effects of core components.}
Compared with random selection, knowledge-gap selection improves
distillation-training and distillation-test accuracy by $9.7$ and $6.0$
percentage points, respectively, maintains the same retention accuracy,
and yields one additional correct response on the frontier subset.
These results suggest that selecting synthesis problems according to
the Teacher--\textit{Student} knowledge gap provides more targeted
teaching signals and a more consistent overall improvement.

Removing the Prompter yields the highest distillation-training accuracy
of $72.4\%$, but reduces distillation-test and retention accuracy from
$56.0\%$ and $72.5\%$ to $48.8\%$ and $70.0\%$, respectively, while
frontier accuracy remains at $5.0\%$.
This trade-off suggests that direct \textit{Teacher} demonstrations fit
the synthesis problems more strongly, whereas the Prompter helps transfer
the acquired knowledge to problems not used for synthesis.

Without validation gating, distillation-training, distillation-test,
and frontier accuracy decrease from $65.3\%$, $56.0\%$, and $5.0\%$
to $59.2\%$, $39.3\%$, and $3.8\%$, respectively, while retention
accuracy increases from $72.5\%$ to $80.0\%$.
Thus, removing the gate does not yield a consistent improvement, but
instead trades lower accuracy on the other three subsets for higher
retention accuracy.

\textbf{Effect of batch size.}
Among the three batch sizes, the default batch size of 16 performs best
on the distillation-training and distillation-test subsets and ties
batch size 32 on the retention and frontier subsets.
The smaller batch size of 8 performs worse across all four subsets.
Although batch size 32 maintains the retention and frontier results,
its distillation-training and distillation-test accuracy decreases to
$59.2\%$ and $45.2\%$, respectively.
These results suggest that an intermediate batch size provides a better
balance between the amount of teaching evidence consolidated in each
update and the frequency of \textit{Primer} refinement.
\subsection{Effects of Quantization and Thinking Mode}

\begin{table}[!ht]
\centering
\caption{\textit{Primer} gains under different quantization and thinking settings of Qwen3.6-27B on KernelBench.}
\label{tab:reasoning-quantization-settings}
{\footnotesize
\setlength{\tabcolsep}{3pt}
\renewcommand{\arraystretch}{0.95}
\setlength{\aboverulesep}{0.35ex}
\setlength{\belowrulesep}{0.45ex}

\begin{tabular*}{\linewidth}{@{\extracolsep{\fill}}lcccccc@{}}
\toprule
\multirow{2}{*}{\textbf{Configuration}}
& \multicolumn{3}{c}{\textbf{Accuracy (\%)}}
& \multicolumn{3}{c}{\textbf{Fast$_1$ (\%)}} \\
\cmidrule(lr){2-4}\cmidrule(lr){5-7}
& Baseline & With \textit{Primer} & Gain
& Baseline & With \textit{Primer} & Gain \\
\midrule
Unquantized
& 9.2 & \textbf{41.2} & \textbf{+32.0}
& 12 & \textbf{37} & \textbf{+25} \\
\midrule
Int4 RTN (G128)
& 4.4 & 34.6 & +30.2
& 5 & 28 & +23 \\
Int4 RTN (Channelwise)
& 0.4 & 21.6 & +21.2
& 0 & 16 & +16 \\
\midrule
Thinking disabled
& 1.6 & 16.6 & +15.0
& 1 & 7 & +6 \\
\bottomrule
\end{tabular*}

\par\smallskip
\noindent
\begin{minipage}{\linewidth}
\footnotesize
\setlength{\parskip}{0pt}
\textit{Note.} The \textit{Primer} is synthesized under Pro$\rightarrow$Flash and transferred unchanged to all Qwen configurations.
RTN denotes round-to-nearest quantization; G128 denotes a group size of 128, and Channelwise denotes per-channel quantization.
Thinking is enabled in the first three configurations, while the final configuration uses the unquantized model with thinking disabled.
Gains are measured in percentage points, and bold marks the best primed result and gain for each metric.
\end{minipage}
}
\end{table}

Table~\ref{tab:reasoning-quantization-settings} evaluates the same transferred \textit{Primer} across unquantized Qwen3.6-27B with thinking enabled, two Int4 RTN configurations \citep{rtn}, and the unquantized model with thinking disabled.
The \textit{Primer} improves both accuracy and Fast$_1$ in all four settings.
Although the gain magnitude varies across configurations, the same \textit{Primer} remains effective without resynthesis under the evaluated changes to quantization and thinking mode.
\section{Primer Analysis and Examples}
\label{app:primer-examples}

This section presents \textit{Primer} examples for mathematical reasoning (Omni-MATH-2), GPU kernel generation (KernelBench), and multimodal geometry reasoning (MathVerse) to illustrate the content and organization of the text synthesized by UTT.
These \textit{Primers} contain task knowledge, problem-solving strategies, and guidance for avoiding failures, with different emphases depending on the task.
The following \textit{Primers} preserve the original model-generated content
without factual correction; only Markdown and LaTeX formatting is
normalized. They may contain inaccurate, incompletely qualified, or
problem-specific statements.
We first summarize the content and characteristics of each example, followed by the full \textit{Primer} texts.

\textbf{Mathematical reasoning Primer.}
The \textit{Primer} in Example~\ref{ex:primer-math} organizes problem-solving
records from Olympiad problems by topic, covering geometry, number
theory, polynomial values, graph-based constructions, combinatorial
arguments, and probability.
In addition to specific constructions and derivations, it records
validation practices such as checking boundary cases, verifying explicit
constructions, and testing potential counterexamples.
This example preserves the compact form produced by the \textit{Synthesizer} and
illustrates how problem-solving experience from different problems is
consolidated into a textual \textit{Primer}.

\textbf{GPU kernel generation Primer.}
The \textit{Primer} in Example~\ref{ex:primer-kernel} emphasizes output-format
compliance and the basic requirements for custom CUDA implementations,
and summarizes extension binding, scan-style operators, lazy compilation,
and common compilation and runtime errors.
This example illustrates how a \textit{Primer} organizes code-generation
requirements, implementation patterns, and engineering troubleshooting
experience for CUDA operators into textual guidance that can be directly
used by the \textit{Student}.
This example illustrates how a \textit{Primer} combines computational patterns, implementation details, and engineering experience to guide code generation and validation.

\textbf{Multimodal geometry reasoning Primer.}
The \textit{Primer} in Example~\ref{ex:primer-mathverse} organizes figure interpretation, geometric reasoning, and answer output into a sequential problem-solving process.
It guides the \textit{Student} to associate numerical labels with geometric objects, distinguish easily confused concepts such as radius and diameter, and solve for the target quantity using tools such as angle propagation, similarity relations, and area decomposition.
It also addresses repeated figure interpretation and reasoning budget exhaustion through strategies that limit alternative interpretations, focus on the target quantity, and ensure timely answer output, illustrating concrete forms of textual guidance for multimodal problem solving.

\begin{UTTPrimerBox}[label={ex:primer-math}]
{Mathematical Reasoning (DeepSeek V4 Pro $\rightarrow$ DeepSeek V4 Flash)}

\textbf{Primer: Olympiad Problem-Solving Patterns}

\UTThead{0 Output discipline}

\begin{itemize}
\item End with a finite checkable proof/boxed answer; prefer compact algebra,
counting, or graph arguments.

\item Verify definitions, boundary cases, and small examples before trusting
analogies or known results.
\end{itemize}

\UTThead{1 Geometry}

\begin{itemize}
\item Circle tangency via coordinates: normalize; circle through points
\(x^2+y^2+ax+by+c=0\); prove common point and proportional gradients.

\item Billiards by unfolding: reflect polygon, not ray; path becomes straight
segment in tiling; use gcd/color conditions for vertex hits.

\item Distinguish metric vs combinatorial definitions; use explicit
polar/nonregular counterexamples.

\item Tetrahedron altitude from six edges: build base \(B'C'D'\); build
rotated face points \(A_B\) opposite side of \(C'D'\) with
\(A_BC'=AC, A_BD'=AD\), and \(A_C\) opposite side of \(B'D'\) with
\(A_CB'=AB, A_CD'=AD\). Perpendiculars from \(A_B\) to line \(C'D'\)
and \(A_C\) to line \(B'D'\) meet at the true foot \(X'\). Altitude
\(=\sqrt{|A_BY'|^2-|X'Y'|^2}\), constructible by Pythagorean difference.
Use full supporting lines, not segments. Foot from \(A\) to line \(BC\)
has signed \(BP=(AB^2+BC^2-AC^2)/(2BC)\); \(BP<0\) or \(BP>BC\) means
foot outside segment. Transfer signed positions; unsigned distances fail
in obtuse cases.
\end{itemize}

\UTThead{2 Number theory and polynomial values}

\UTThead{2.1 Vieta jumping/descent}

For \((a+b)(a+b+1)/(ab)=N\): fix \(N\), rewrite as quadratic, Vieta
gives positive integer other root; descend to equal pair; verify
integrality/positivity.

\UTThead{2.2 Lyndon words and fractional parts}

Suffix/prefix lex conditions become Lyndon words; count length \(L\)
over \(q\) by
\(\frac1L\sum_{d\mid L}\mu(d)q^{L/d}\).

\UTThead{2.3 Reduced denominators / averaging}

For \(S_n=A_n/n!\), denominator \(n!/\gcd(A_n,n!)\); lifts
\(A_{r+p^k}\equiv A_r-p^k\pmod{p^{k+1}}\); existence via averaging
and Stirling.

\UTThead{2.4 Sum-of-\(k\)-others and omitted elements}

If each element is a sum of \(k\) others in a set of \(k+m\), count
omitted elements; compare largest/smallest omitted sums. Don't use
sign-count bounds: largest need not be positive sum. For zero-sum
symmetric sets solve \(a+b+c=-2x\).

\UTThead{2.5 Digit sums and carries}

\(\sigma(10m+i)=\sigma(m)+i\); safe block iff
\(\sigma(m)\equiv1\pmod{11}\); trailing-9s give
\(\sigma(m+1)-\sigma(m)=1-9t\). Maximal safe intervals shape
\(9+10+10+9\).

\UTThead{2.6 Integer-valued polynomials and \(P(i)\)}

\begin{itemize}
\item Integer-valued iff
\(P(x)=\sum c_k\binom{x}{k}\), \(c_k\in\mathbb Z\).

\item At \(x=i\), analyze
\(v_\pi(\binom{i}{k})=v_\pi(N_k/k!)\).
For \(p\equiv1\pmod4\), \(p=\pi\bar\pi\) in \(\mathbb Z[i]\);
Hensel gives \(s\) with \(\pi^e\mid i-s\), so among
\(i,\dots,i-k+1\) at least \(\lfloor k/p^e\rfloor\) are divisible by
\(\pi^e\); hence no \(p\equiv1\pmod4\) appears in denominators.

\item \(p=2\) or \(p\equiv3\pmod4\) attainable:
\(1/2=6\binom{x}{4}+3\);
\(1/p=(p-1)!\binom{x}{p}\).

\item Answer: \(a+bi\), \(a,b\in\mathbb Q\), with
\(\nu_p(a),\nu_p(b)\ge0\) for every \(p\equiv1\pmod4\).
Not all \(\mathbb Q(i)\): \(1/5\) excluded.
\end{itemize}

\UTThead{2.7 Prime-divisor constraints \(\omega(n)>K\)}

\begin{itemize}
\item Strict \(>K\) means minimal \(\omega(n)=K+1\); constants \(c\)
need \(c>0\), \(\omega(c)\le K+1\). Boundary \(\omega(c)=K+2\) with
\(\omega(n)=K+1\).

\item Monomials \(x^m\) work.

\item Nonconstant nonmonomials fail. If \(P(0)\ne0\), Schur gives
\(K+2\) primes \(q_i\) with nonzero roots \(a_i\bmod q_i\);
CRT+Dirichlet builds \(n\) with exactly \(K+1\) prime factors and
\(n\equiv a_i\bmod q_i\); then \(q_i\mid P(n)\), \(q_i\nmid n\), so
\(\omega(P(n))\ge K+2\) or \(P(n)\le0\). If \(P(0)=0\), factor
\(x^mQ(x)\), \(Q(0)\ne0\), apply to \(Q\); handle constant factor
\(c>1\).
\end{itemize}

\UTThead{3 Scheduling and exact-degree constructions}

\begin{itemize}
\item Tournament stays: interval per player; pairwise intersection + Helly
gives common day; one match at central day; total cost baseline \(2M\)
plus idle player-days; balance distinct players before/after.

\item Touch/degree: for \(n\) objects each touching exactly 3 others,
handshake gives \(3n\) even, so \(n\) even; do not infer divisibility
by 4. For large even \(n\), cycle base (outside squares on \(4m\)-gon
plus reflected pairs gives \(6m\)) and insert four-square gadgets at
cuts to add 8 or 16, covering residues mod 6; verify no unintended
intersections.
\end{itemize}

\UTThead{4 Hamiltonian paths on grids}

\begin{itemize}
\item Numbering an \(n\times n\) grid is a Hamiltonian path; diagonal
projection gives a \(\pm1\)-walk, main diagonal visits same parity.
Bound first/last visits using side color counts; construct explicitly.
\end{itemize}

\UTThead{5 Permutation reachability via swap graphs}

\begin{itemize}
\item Allowed swaps = graph edges; connected graph iff every permutation
reachable. Prove connectivity by explicit spanning path; disprove by
separated component.
\end{itemize}

\UTThead{6 Local block constraints and two-stage counting}

\begin{itemize}
\item Count distinguished positions first, then assign remaining; encode
as \(0,\pm1\). For \(2\times2\) zero-sum blocks, general solution
\(a_{ij}=(-1)^{i+j}(r_i+c_j)\); propagate sign choices across overlaps.
\end{itemize}

\UTThead{7 Hyperplane-generic finite sets}

\begin{itemize}
\item Minimal \(k\)-generic sets: lower via concurrent lines with \(k\)
points; upper via private-point subcover and affine linear functions;
incidence/basis gives \(|M|\le kn\) for \(k,n>1\).
\end{itemize}

\UTThead{8 Partition minima and smoothing}

\begin{itemize}
\item Define
\(S(N)=\min_{n_1+\cdots+n_m=N}\sum f(n_i)\) for nonincreasing \(f\).
Balanced partition is only an upper bound:
\(S(N)\le r f(q+1)+(m-r)f(q)\), \(N=mq+r\), \(0\le r<m\).
Summing one fixed partition family overcounts because \(S\) is the
minimum over all partitions. Count all zero parts: for \(m=20\),
\(N=0,\dots,19\), coefficient of \(f(0)\) is \(20+\cdots+1=210\),
not 20.

\item Smoothing extremal: if
\(\sum f(i)=A=\binom{a+1}{2}\), then
\(\sum_{N\ge0}S(N)\le
\sum_{N=0}^{ma}(ma-N)=ma(ma+1)/2\).
Equality at \(f(i)=\max(a-i,0)\), \(g(N)=\max(ma-N,0)\).
Verify: \(f(n_i)\ge a-n_i\), so sum \(\ge ma-N\), and sum \(\ge0\).
\end{itemize}

\UTThead{9 Fair bounded selection with biased coins}

\begin{itemize}
\item Do not assume arbitrary outcome probabilities can be partitioned
into \(n\) equal groups. For \(n=3\), \(p=1/3\), \(L=3\), weights
\(8,4,4,4,2,2,2,1\) cannot split into three sums 9
(8 must pair with 1; remaining even weights cannot sum 9).

\item Verified construction: choose \(p_1=1/n\), \(p_2=1/2\); take
\(c\) with \(N=2^c\ge n(n-1)\), announce \(L=c+1\). Toss coin1 once,
coin2 \(c\) times. In units \(1/(nN)\), there are \(N\) \(T\)-outcomes
of weight \(n-1\) and \(N\) \(H\)-outcomes of weight 1; each person
needs total weight \(N\). Write \(N=qn+r\), \(0\le r<n\); set
\(a_i=q+1\) for \(r\) people, \(a_i=q\) otherwise;
\(b_i=N-a_i(n-1)\). Feasibility \(a_i(n-1)\le N\) follows from
\(N\ge n(n-1)\). Assign \(a_i\) \(T\)-strings and \(b_i\)
\(H\)-strings to person \(i\); probability
\(a_i(n-1)/(nN)+b_i/(nN)=1/n\). Explicitly verify the partition before
announcing.
\end{itemize}

\UTThead{10 Validation checklist}

\begin{itemize}
\item Integer-valued polynomials: binomial basis; check split primes
\(p\equiv1\pmod4\) separately.

\item \(\omega(n)>K\): convert to \(K+1\); constants separately;
Schur+CRT/Dirichlet to force prime divisors.

\item Touch/degree: handshake parity; avoid extra divisibility; test
constructions for unintended coincidences.

\item Scheduling/grid/digit/local-counting: include idle days; diagonal
parity and explicit attainment; carry blocks; propagate overlap
consistency.

\item Vieta jumping: verify positive integral new root. Circle tangency:
check membership and gradients. Counting formulas: check
divisors/M\"obius signs.

\item Tetrahedron altitude: use supporting lines and signed foot distances;
test obtuse cases (\(BP<0\) or \(BP>BC\)); verify intersection is true
projection.

\item Partition minima: balanced partition is upper bound not exact;
count zero coefficients; verify extremal construction.

\item Fair selection: verify explicit equiprobable partition for announced
bound; test small \(n\)/parity counterexamples.
\end{itemize}

\end{UTTPrimerBox}

\begin{UTTPrimerBox}[label={ex:primer-kernel}]
{GPU Kernel Generation (DeepSeek V4 Pro $\rightarrow$ DeepSeek V4 Flash)}

\textbf{PRIMER: Writing Custom CUDA Operators for PyTorch Models}

\UTThead{Core Requirements}

\begin{itemize}
\item Replace operators with a \textbf{custom CUDA implementation},
typically inside a \texttt{torch.autograd.Function} or as a direct
extension function.

\item \textbf{No} \texttt{try}/\texttt{except} or fallback logic ---
let assertions crash.

\item \textbf{Output format is critical:} the entire answer must be raw
Python source, starting with the first character and continuing until
the end. If the instruction says ``output only the code,'' the response
must be solely the code block---no introductory or trailing text and no
Markdown fences.

\item All methods must be fully implemented, with no placeholders.
\end{itemize}

\UTThead{Token Budget \& Code-First Strategy}

\begin{itemize}
\item \textbf{Code output must come immediately.}
Spending the budget on reasoning before the code can lead to truncation,
leaving no answer at all.

\item Prefer a \textbf{minimal, correct kernel}, such as one thread per
output element or a serial scan, to remain within the token limit.

\item For inference-only tasks, skip backward computation by raising
\texttt{NotImplementedError}.

\item Target roughly 60--80 lines of model code; simplify kernels that
grow substantially larger.
\end{itemize}

\UTThead{Symbol Visibility Before Binding (Critical)}

Every function referenced by \texttt{m.def} must be known at the point
of registration. The extension binding code is a plain C++ translation
unit and does not permit linking later to unresolved symbols. Two safe
patterns are:

\begin{enumerate}
\item \textbf{Monolithic source (recommended):}
define all functions before the \texttt{PYBIND11\_MODULE} block inside
a single CUDA source string and leave \texttt{cpp\_sources=[]}.

\item \textbf{Multi-file:}
declare the wrapper function in a header included by the binding file.
Its definition in a separate \texttt{.cu} file must exactly match the
declaration.
\end{enumerate}

The host function signature must be compatible with
\texttt{torch::wrap\_pybind\_function}, conceptually a
\texttt{std::function} accepting the specified argument types and
returning a \texttt{torch::Tensor}. Using \texttt{\&my\_func} directly
is simpler and preferred.

\UTThead{Monolithic Template (Safe)}

\begin{Verbatim}[
  fontsize=\scriptsize,
  breaklines=true,
  breakanywhere=true,
  breaksymbolleft={}
]
cuda_src = r'''
#include <torch/extension.h>
__global__ void my_kernel(...) { ... }
void my_op(torch::Tensor x, torch::Tensor y) { ... /* launch kernel */ }
PYBIND11_MODULE(TORCH_EXTENSION_NAME, m) {
  m.def("my_op", &my_op, "doc");
}
'''
ext = load_inline(name="my_ext", cpp_sources=[],
                  cuda_sources=[cuda_src], verbose=False)
\end{Verbatim}

\UTThead{General Kernel Design Patterns}

\begin{itemize}
\item \textbf{One thread per output} for convolution, pooling, and
similar operations: launch one thread per output element, loop over
kernel and channel dimensions, and use \texttt{\_\_ldg} for read-only
data.

\item \textbf{Scan-type operations} such as cumulative sums:
default to one thread per independent slice, looping sequentially over
the scan dimension. This is simple, token-efficient, and reliable.

\item \textbf{Reverse cumulative sums:}
loop from right to left rather than composing flip, cumulative sum,
and another flip.

\item \textbf{Multidimensional indexing:}
parenthesize expressions aggressively to avoid compilation failures
caused by missing parentheses.
\end{itemize}

\UTThead{Implementing Scan Operations}

\begin{itemize}
\item Treat the tensor as \texttt{(num\_vectors, L)}, moving the target
dimension to the final position when necessary.

\item Launch \texttt{num\_vectors} threads, each performing a serial
inclusive scan.

\item For a reverse cumulative sum, iterate from right to left.

\item Let the host wrapper move the target dimension to the end, launch
on a two-dimensional view, and permute the result back.
\end{itemize}

\UTThead{Extension Building with \texttt{load\_inline}}

\begin{itemize}
\item Exactly one \texttt{PYBIND11\_MODULE} must appear, with
\texttt{TORCH\_EXTENSION\_NAME} as its first argument.

\item The \texttt{name} supplied to \texttt{load\_inline} must match
the module name represented by the macro.

\item Use \texttt{extra\_cflags=["-O3"]} and
\texttt{extra\_cuda\_cflags=["-O3"]}.

\item Compile once and cache the module at the class level to avoid
recompilation for every instance.
\end{itemize}

\UTThead{Lazy Compilation (Class-Level Caching)}

\begin{Verbatim}[
  fontsize=\scriptsize,
  breaklines=true,
  breakanywhere=true,
  breaksymbolleft={}
]
class ModelNew(nn.Module):
    _ext = None
    def __init__(self, dim):
        if ModelNew._ext is None:
            ModelNew._ext = load_inline(
                name="op",
                cpp_sources=[],
                cuda_sources=[cuda_src],
                verbose=False)
        self.ext = ModelNew._ext
\end{Verbatim}

\UTThead{Testing Builds Locally}

Verify that a small extension compiles and executes before integrating it:

\begin{Verbatim}[
  fontsize=\scriptsize,
  breaklines=true,
  breakanywhere=true,
  breaksymbolleft={}
]
ext = load_inline(name="test", cpp_sources=[],
                  cuda_sources=[cuda_src], verbose=True)
x = torch.randn(1, 1, 8, 8, 8).cuda()
ext.my_op(x, ...)
\end{Verbatim}

\UTThead{Common Failure Modes}

\begingroup
\setlength{\tabcolsep}{4pt}
\renewcommand{\arraystretch}{1.12}
\setlength{\parskip}{0pt}

\begin{UTTrow}
\hline
\textbf{Symptom} & \textbf{Root cause and fix} \\
\hline
\end{UTTrow}

\begin{UTTrow}
\texttt{SyntaxError} at the start of the compiled string
& Markdown fences or leading text were included; output pure Python
source. \\
\end{UTTrow}

\begin{UTTrow}
No code output; token budget exhausted
& The model spent the budget on reasoning; emit code immediately. \\
\end{UTTrow}

\begin{UTTrow}
\texttt{redefinition of PyInit\_xxx}
& Multiple \texttt{PYBIND11\_MODULE} blocks; retain exactly one. \\
\end{UTTrow}

\begin{UTTrow}
\texttt{my\_func was not declared}
& The function is not visible before \texttt{m.def}; define it first
or include its declaration. \\
\end{UTTrow}

\begin{UTTrow}
\texttt{TypeError: SupportsFloat}
& The host function expects raw pointers; accept
\texttt{torch::Tensor} and extract pointers internally. \\
\end{UTTrow}

\begin{UTTrow}
Wrong output or segmentation fault
& Non-contiguous strides were ignored; call
\texttt{.contiguous()} or handle strides explicitly. \\
\end{UTTrow}

\begin{UTTrow}
CUDA expected a semicolon
& A complex index expression has unbalanced parentheses; introduce
intermediate variables and balance the expression. \\
\end{UTTrow}

\begin{UTTrow}
Extension-name mismatch
& The supplied \texttt{name} and binding module differ; use
\texttt{PYBIND11\_MODULE(TORCH\_EXTENSION\_NAME, ...)}. \\
\hline
\end{UTTrow}

\endgroup

\UTThead{Verification Checklist}

\begin{itemize}
\item Output is pure Python source, without fences or commentary when
these are forbidden.

\item Code is emitted immediately without prolonged reasoning.

\item Exactly one \texttt{PYBIND11\_MODULE} is present, and every
referenced host function is visible beforehand.

\item Host functions accept PyTorch-compatible types rather than raw
pointers.

\item No fallback logic appears in the optimized implementation.

\item Tensor contiguity is checked or handled.

\item Reverse scans traverse in reverse rather than composing flips.

\item Multidimensional index expressions are balanced and tested.

\item The extension is tested on a small tensor before a full benchmark
run.
\end{itemize}

\end{UTTPrimerBox}

\begin{UTTPrimerBox}[label={ex:primer-mathverse}]{Multimodal geometry reasoning Primer (Claude Opus 5 $\rightarrow$ Qwen 3.6 27B)}

\textbf{PRIMER: Multiple-Choice Geometry from a Figure}

\UTThead{0. Non-negotiable output rule}

One letter is the deliverable.
\textbf{Write "Answer: X" first
(or end with the bare letter on its own line),
then 2--6 lines of justification.}
Nearly every observed failure was identical:
\emph{no final output} ---
budget burned re-reading the figure,
cataloguing every labeled value,
or debating which vertex owns a number.
Partial reasoning scores zero; a best guess always beats silence.

Budget policy: 2--4 reasoning steps,
\textbf{at most one or two figure interpretations}, then stop.

\begin{itemize}
\item Value matches an option $\rightarrow$ commit; stop verifying.

\item Matches nothing $\rightarrow$ say so in one sentence,
switch the assignment once, else pick the \textbf{nearest listed value}.

\item Never solve for every lettered unknown;
stop the instant the target is pinned.

\item "Cannot be determined" only after a determinate chain visibly
breaks --- never as a hedge.
\end{itemize}

\UTThead{1. Read the figure, then isolate one unknown}

\begin{itemize}
\item Attach each number to the vertex/side/bracket its mark touches;
name it ("$\angle MLJ=30^\circ$").
An arrow from a value just labels the adjacent lettered angle
($r^\circ\leftarrow90$ means $r=90$).

\item Decide \textbf{what each length denotes}:
radius vs diameter, whole bracket vs sub-segment,
side vs apothem, slant vs height.
Label-swapping is the \#1 distractor generator.
An arrow drawn inside a small circle from its center
is that circle's radius.

\item List only the structural facts you need:
which point-triples are collinear,
which segments are parallel/tangent.

\item \textbf{Chain, don't systematize:}
anchor on the triangle/relation with two knowns,
propagate, stop.

\item \textbf{Prune decoys explicitly};
extra labels often serve other sub-questions.

\item Not-to-scale figures: test candidate readings against options;
keep the clean one.
\end{itemize}

\UTThead{2. Angle propagation toolkit}

Two tools carry most crossing-line figures:
\textbf{vertical angles} at the intersection and
\textbf{$180^\circ$ sums} (triangle, straight angle).
Template: triangle with two given angles
$\rightarrow$ third angle
$\rightarrow$ vertical angle across
$\rightarrow$ next triangle
$\rightarrow$ straight-angle subtraction
$\rightarrow$ target.

Parallel lines: arrowheads declare the parallel pair;
find the transversal joining target to given.
Corresponding/alternate interior $\Rightarrow$ equal;
co-interior or linear pair $\Rightarrow$ supplementary.
Misreading same-side as alternate gives the $180-\theta$ distractor.
Extended ray $\Rightarrow$ linear pair.
In a parallelogram a diagonal is a transversal for both side pairs;
pair the correct halves.
Quadrilateral interior sum $360^\circ$.

\UTThead{3. Congruent/joined figures}

Congruent polygons glued along a shared side:
corresponding angles transfer;
the angle at a seam vertex is usually the \textbf{sum}
of one angle from each.
Validate with polygon angle sums.

\UTThead{4. Similar triangles (parallels, shadows, mirrors)}

$AB\parallel CD$ with apex $P$
$\Rightarrow\triangle PAB\sim\triangle PCD$;
$h(P\rightarrow AB)=(AB/CD)\cdot h(P\rightarrow CD)$.
Distance between parallels $=h_{\mathrm{total}}-h_{\mathrm{sub}}$
(must be $<$ total).
Mirror/shadow:
$h_{\mathrm{far}}/h_{\mathrm{near}}
=d_{\mathrm{far}}/d_{\mathrm{near}}$.

\UTThead{5. Circle toolkit}

\begin{itemize}
\item \textbf{Find diameters first:}
collinear labeled points through the center
$\Rightarrow$ adjacent central angles are a linear pair;
all central angles sum 360.

\item \textbf{Arc notation:}
two letters = minor arc = central angle;
\textbf{three letters = arc through the middle letter},
usually major $\Rightarrow360-\mathrm{minor}$ ($>180^\circ$).

\item Inscribed = $\frac12$ arc;
central = $2\times$inscribed;
cyclic quad opposite angles supplementary;
chord = $2R\sin A$.

\item Tangent from external $P$:
tangent $\perp$ radius;
$AP=OP\cdot\sin\angle AOP$,
$r=OP\cdot\cos\angle AOP$;
check $r<OP$.

\item Equidistant chords $\Rightarrow$ congruent chords
$\Rightarrow$ congruent arcs
$\Rightarrow$ equal inscribed angles.

\item Internally tangent circles:
centers and contact point are collinear.
Two equal circles of radius $r$ internally tangent to a big circle
at opposite ends of a diameter and tangent to each other
$\Rightarrow R=2r$.

\item \textbf{Perimeter of a curvilinear region = sum of arc lengths,
each $(\theta/360)\cdot2\pi\rho$ with its own radius.}
Half of a circle of radius $\rho$ contributes $\pi\rho$.
Typical shaded lune: $\pi R+\pi r+\pi r$.
Rectangle on a semicircle diameter:
$h=\sqrt{R^2-(w/2)^2}$.
\end{itemize}

\UTThead{6. Area by symmetry and cancellation}

Shaded = whole $-$ unshaded; hunt equal-area pieces first.
\textbf{Common apex $\Rightarrow$ compare triangles by base alone:}
apex at a rectangle's center gives every top/bottom triangle
height $H/2$, area $bH/4$,
so group bases until they sum to a full side
(top set = bottom set = $\frac14$ of area; sides = $\frac12$).
If a shaded base and its mirror unshaded base are equal,
shaded bases total one full width
$\Rightarrow\frac14$ of the rectangle.
Verify all region fractions sum to 1.
Assume convenient dimensions ($4\times2$) or coordinates.
Quarter disc radius $r$ and semicircle on its chord
both have area $\pi r^2/4$
$\Rightarrow$ shaded = $\frac12r^2$;
$\pi$-free answers signal cancellation.

\UTThead{7. Cone/net toolkit}

$(\theta/360)\cdot2\pi R=2\pi r$;
$\theta=90^\circ\Rightarrow R=4r$.
Lateral surface = $\pi rl$;
$l^2=h^2+r^2$.
$AE=AF$, $DE=DF\Rightarrow$ axis $\perp$ base.

\UTThead{8. One-line validation, then emit}

Supplements sum 180; polygon sums close;
drawn obtuse $\Rightarrow>90^\circ$;
chord $\leq$ diameter;
$r<$ distance to external point;
fractions sum to 1;
magnitude plausible against options.
Name the licensing relation per step.
Then emit the letter.

\end{UTTPrimerBox}
\section{Limitation Discussion and Future Work}\label{app:limitation_future_work}
We are aware that UTT increases inference-time token usage because the \textit{Primer} is prepended to every task input, which could be considered a limitation of the method. Future work can reduce this overhead through \textit{Primer} compression or retrieval-based selection of relevant content. 

Besides, a natural next step of UTT is to distill the knowledge it encodes into the model weights through subsequent training. This direction is most practical for resource-rich organizations with access to model weights and substantial training infrastructure. The primary goal of this work, however, is to provide a proof of concept that knowledge distillation of LLMs can occur in text form, with no weight updates required.

\end{document}